\documentclass[10pt]{article}

\usepackage[T1]{fontenc}
\usepackage{lmodern}
\usepackage{natbib}
\usepackage{comment}
\setcitestyle{authoryear,round,citesep={;},aysep={,},yysep={;}}
\usepackage[margin=1in]{geometry}

\usepackage{amsmath,amsfonts,bm}

\def\eqref#1{equation~\ref{#1}}
\def\1{\bm{1}}

\DeclareMathAlphabet{\mathsfit}{\encodingdefault}{\sfdefault}{m}{sl}
\SetMathAlphabet{\mathsfit}{bold}{\encodingdefault}{\sfdefault}{bx}{n}

\usepackage{hyperref}
\usepackage{url}
\usepackage{graphicx}
\usepackage{algorithm}
\usepackage{algpseudocode}  % or algorithmicx
\usepackage{amsmath}
\usepackage{amssymb}
\usepackage{mathrsfs}
\usepackage{amsthm}
\usepackage{listings}  % for code blocks if needed
\usepackage{xcolor}    % for colored text/code
\usepackage{subcaption}
\usepackage{booktabs}
\usepackage{multirow}
\usepackage{arydshln}
\usepackage{soul}
\usepackage{float}

\usepackage{tikz}
\usetikzlibrary{shapes.geometric, arrows.meta, positioning, fit, backgrounds, calc}

\usepackage{cleveref}

\newcommand{\ourNothreeILnewbestknownbeginN}{82}

\usepackage{authblk}

\title{Geometry-Aware MCTS for Extremal Problems in Combinatorial Geometry}

\author[1]{Luoning Zhang \thanks{luoninz1@uci.edu}}
\author[1]{Xu Zhuang \thanks{xzhuang8@uci.edu}}

\author[1]{Tianhao Wang \thanks{tianhw11@uci.edu }}
\author[1]{Nathan Kaplan \thanks{nckaplan@math.uci.edu}}
\affil[1]{Department of Mathematics, University of California, Irvine}

\begin{document}

\maketitle

\begin{abstract}

We study certain extremal problems in combinatorial geometry that ask about configurations of points in an $n \times n$ grid that satisfy strict, global geometric constraints.  Classical exact solvers suffer from combinatorial explosion for these types of problems, and standard reinforcement learning and transformer-based models struggle with the sparse reward ``validity cliff'' and quadratic token-consumption limits.  To overcome these bottlenecks, we propose a Geometry-Aware Monte Carlo Tree Search (MCTS) framework.  Our approach strictly enforces geometric constraints through incremental updates to the feasible action space.  For constraints about collections of collinear points, like those that occur in the classic No-Three-in-Line problem (Max-N3IL), this mechanism reduces the constraint checking complexity from $O(n^3)$ to $O(n^2)$.  To improve search efficiency, we exploit geometric symmetries in two ways: canonical pruning during node expansion to reduce the branching factor, and symmetric batch transitions to accelerate the discovery of promising configurations. We perform extensive experiments and establish new best-known computational results on five out of six of the problems that we considered.  Notably, for Max-N3IL we find configurations of size roughly $1.8 n$ for grids of size $\ourNothreeILnewbestknownbeginN \le n \le 119$.  For the Smallest Complete Set problem, we find configurations of size roughly $0.95 n$, providing new upper bounds within the tested grids.  This work establishes Geometry-Aware MCTS as a highly adaptable framework for discovering novel configurations in combinatorial geometry.
\end{abstract}

\section{Introduction}

\begin{figure}
    \centering
    \includegraphics[width=0.7\linewidth]{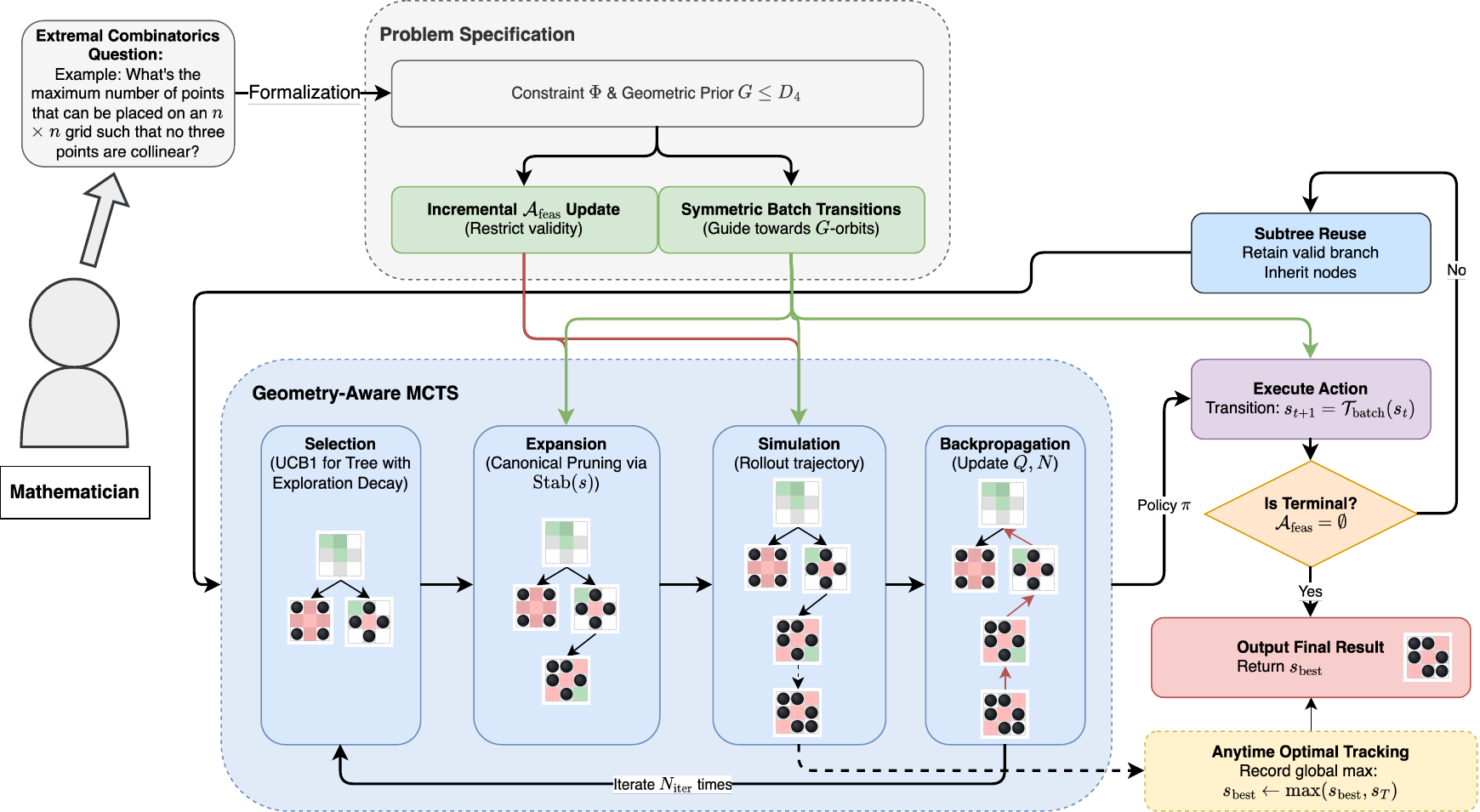}
    \caption{\textbf{Overview of the Geometry-Aware MCTS framework.} The search process is guided by problem specification (grey) including a Feasible Action Space defined by the specific problem constraint and Symmetric Batch Transitions (e.g., rotations and reflections) derived from a geometric prior. The core MCTS engine (blue) utilizes Canonical Pruning to reduce branching during expansion. The resulting policy dictates the main loop (right), which employs Subtree Reuse for efficiency and Anytime Optimal Tracking to asynchronously record the global best-found configuration across all simulations.}
    \label{fig:framework}
\end{figure}

\begin{figure}[ht]
    \centering % Center the entire figure

    % --- First Subfigure ---
    \begin{subfigure}[t]{0.32\textwidth}
        \centering
        \includegraphics[width=0.55\linewidth]{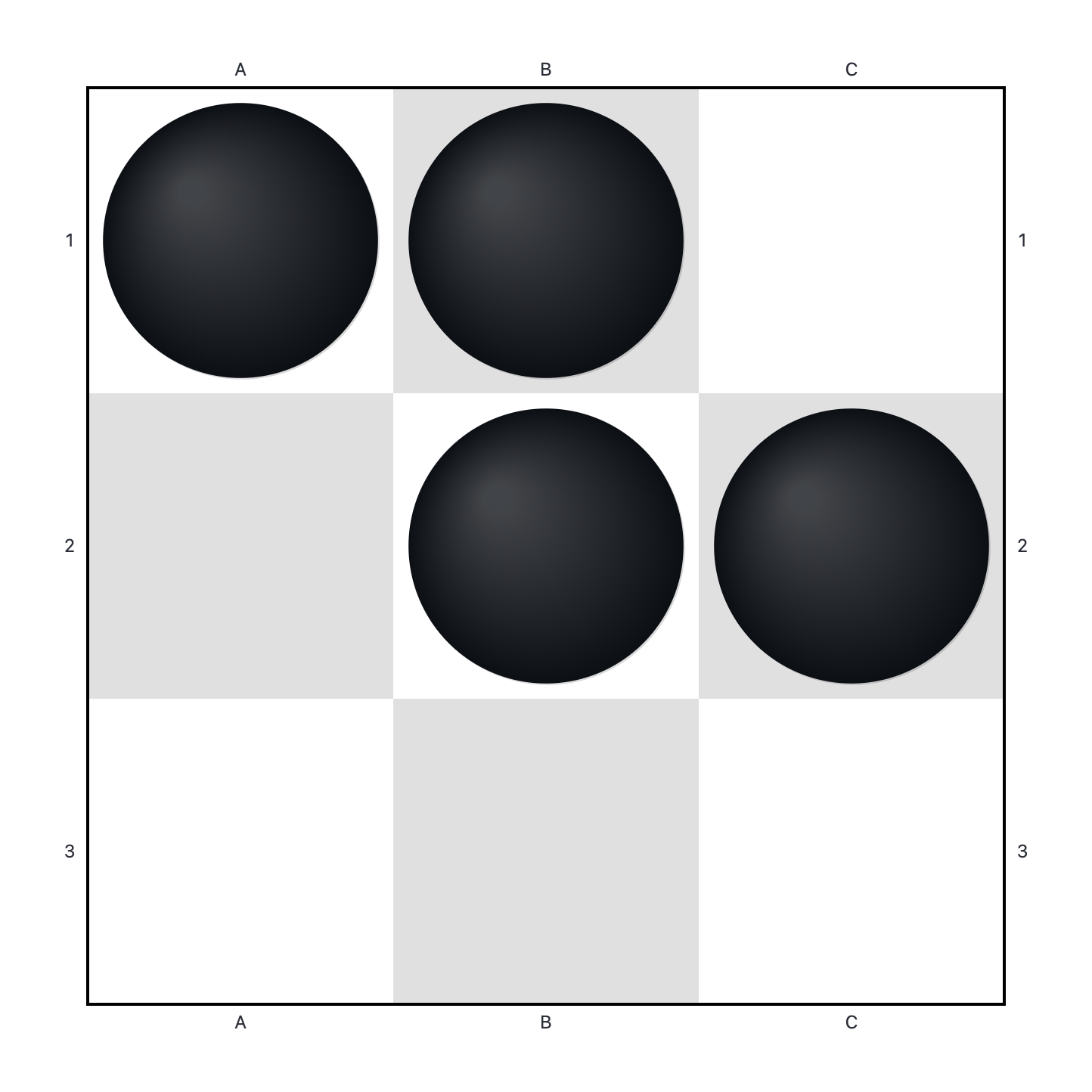}
        \caption{An example configuration with no three collinear points.}
        \label{fig:fig1}
    \end{subfigure}
    \hfill % This command adds horizontal space between the figures
    % --- Second Subfigure ---
    \begin{subfigure}[t]{0.32\textwidth}
        \centering
        \includegraphics[width=0.55\linewidth]{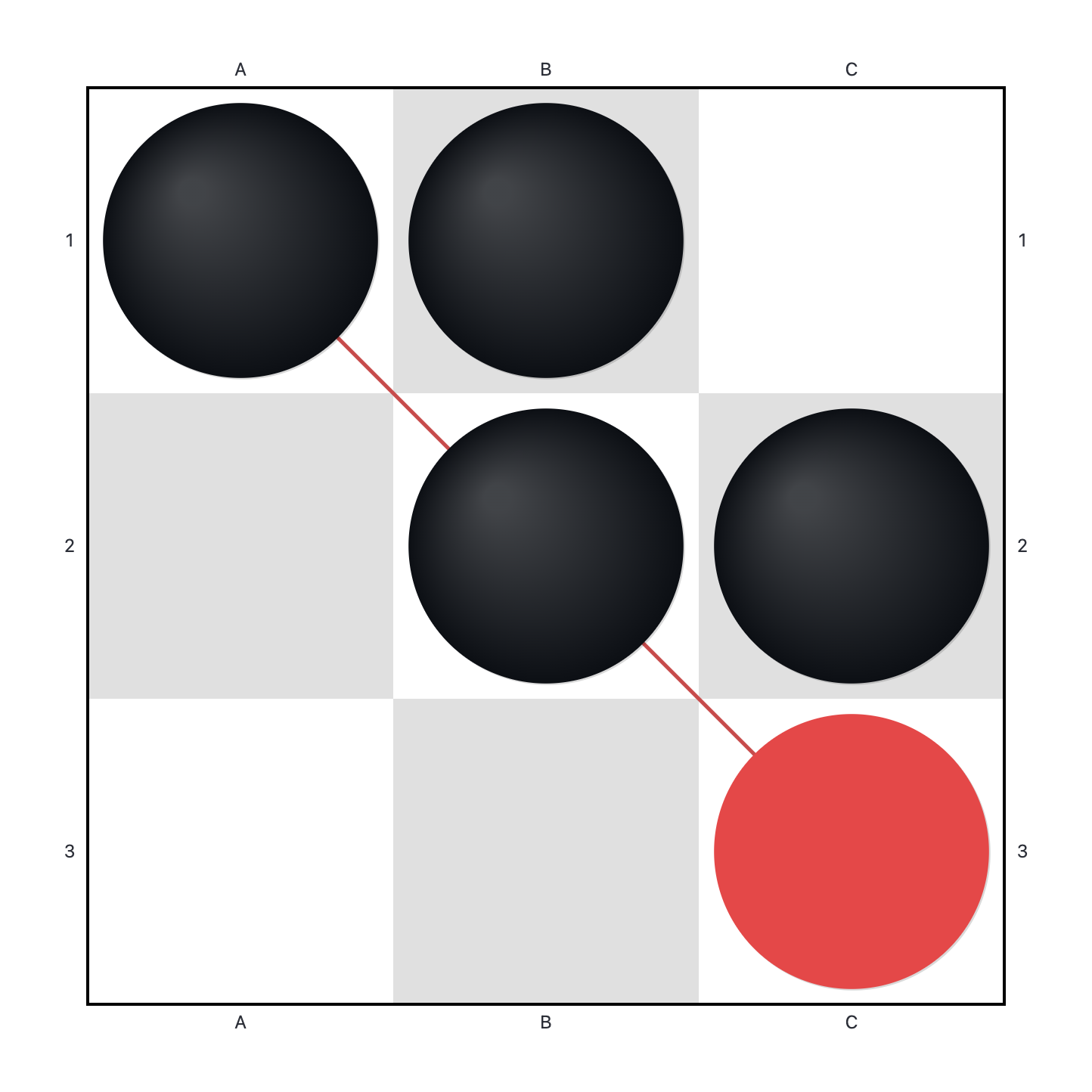}
        \caption{An example configuration with three collinear points (diagonal line).}
        \label{fig:fig2}
    \end{subfigure}
    \hfill % This command adds horizontal space between the figures
    % --- Third Subfigure ---
    \begin{subfigure}[t]{0.32\textwidth}
        \centering
        \includegraphics[width=0.55\linewidth]{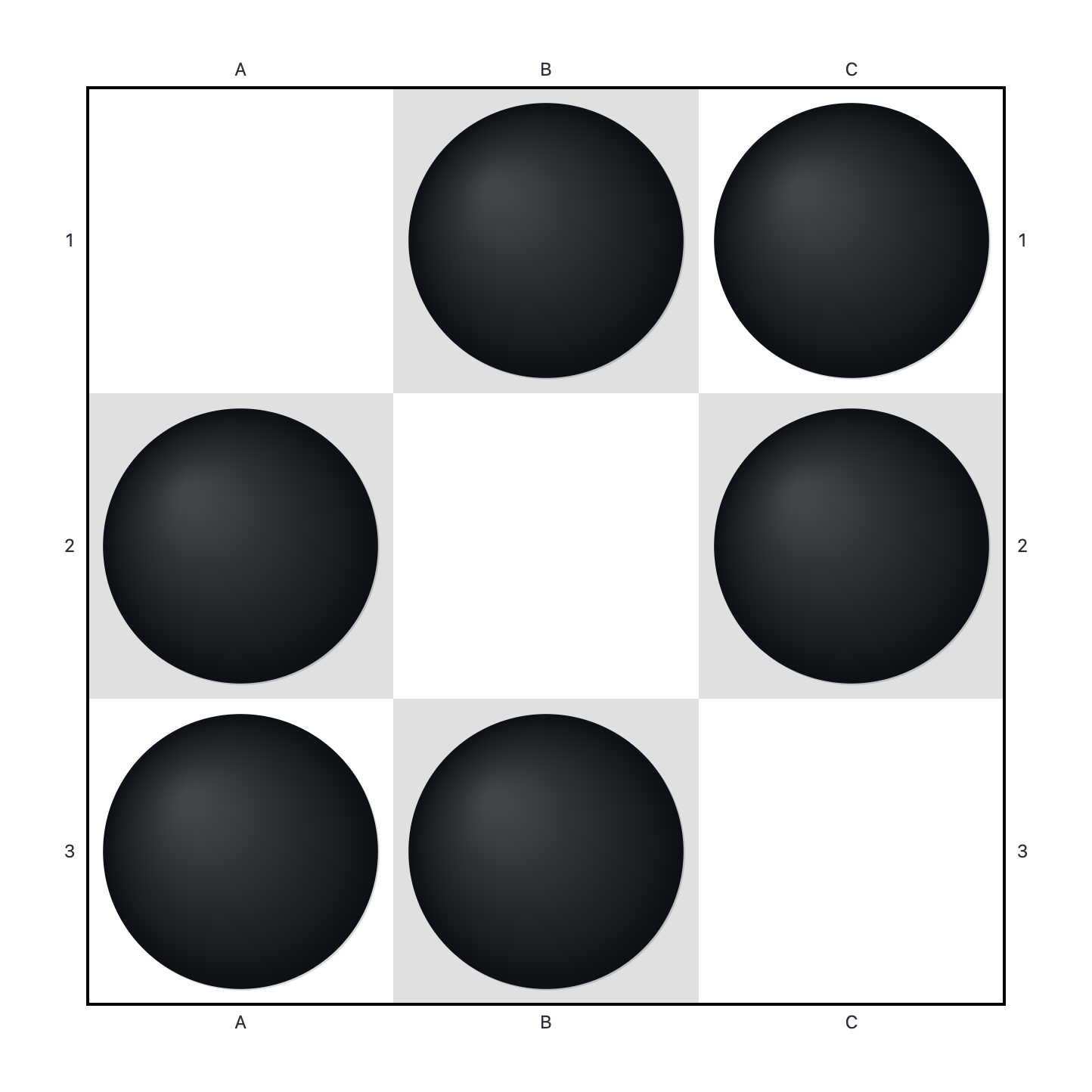}
        \caption{A configuration with no three collinear points of maximum size.}
        \label{fig:fig3}
    \end{subfigure}

    % A single main caption for all the subfigures
    \caption{Three examples of point configurations on a 3x3 grid.}
    \label{fig:three_grids}
\end{figure}

%{\color{red}
%\paragraph{The Problem and the Limitations of Classical Methods}

We study certain extremal problems on $n \times n$ grids.  These problems ask about the maximum or minimum number of points in the grid that satisfy certain geometric constraints.  Representative examples include finding the largest point sets with no isosceles triangles, the largest point sets with no four points on a circle, or the smallest independent geometric dominating set. The No-Three-in-Line Problem (Max-N3IL) has attracted considerable attention, both as a theoretical problem and as an algorithmic one.  This problem has a history going back at least to \citet{dudeney1917}.  It asks for the maximum number of points in an $n \times n$ grid such that no $3$ points are collinear.  We denote the answer to this problem by $M(n)$.  Some configurations in a $3 \times 3$ grid are given in \cref{fig:three_grids}.

Despite its elementary formulation, determining $M(n)$ remains an open challenge for large $n$.  Since $n$ parallel lines cover the $n \times n$ grid and we cannot have more than two points on any of these lines, we see that $M(n) \le 2n$.  \citet{flammenkamp1992, flammenkamp1998} has shown that $M(n) = 2n$ for all $n \le 46$ and for $n \in \{48,50,52\}$.  \citet{prellberg2026constraintsatisfactionprogrammingnothreeinline} has very recently used advanced tools from constraint satisfaction programming to extend these results to show that $M(n) = 2n$ for all $n \le 64$ and for $n \in \{66,68\}$.  Heule used a recently developed SAT solver to find solutions showing that $M(n) = 2n$ for $n \in \{65,67,69, 70\}$.  These results are actively tracked in \citeauthor{flammenkamp_no3in_website}'s online database \citeyearpar{flammenkamp_no3in_website}.  We do not know the exact value of $M(n)$ for any $n \ge 71$.  The best-known asymptotic lower bound is due to \citet{hall1975}, who use a construction coming from finite fields to show that for any $\epsilon > 0$ and for all sufficiently large $n$, $M(n) > (1.5-\epsilon) n$.  This leaves a significant gap between the constructive $(1.5-\epsilon) n$ lower bound and the trivial $2n$ upper bound.  \citet{guy1968} used probabilistic arguments to suggest that as $n$ goes to infinity, $M(n) \approx \frac{\pi}{\sqrt{3}} n \approx 1.814 n$.  In order to better understand the growth of $M(n)$, we would like to probe grid sizes that have historically been inaccessible to computation because of the combinatorial explosion in the search space.

The complexity of these extremal problems on finite grids arises from the global nature of the geometric constraints.  A single point placement affects the set of remaining points in complex ways, making local constraint satisfaction insufficient.  While algebraic constructions provide rigorous guarantees, it is not at all clear whether we should expect extremal configurations to have algebraic structure.  This motivates a need for computational methods to better understand grids that have previously been inaccessible.

In recent years, AI has achieved remarkable breakthroughs in combinatorial mathematics, from discovering graph theory counterexamples \citep{wagner2021} to predicting modular curve invariants \citep{zhuang2025}. Applying these methods has not yet been successful on problems like Max-N3IL. A recent systematic evaluation by \citet{prellberg2025} highlighted severe scalability limits across three distinct computational paradigms: exact solvers, reinforcement learning, and transformer-based models. Traditional exact solvers, such as Integer Linear Programming (ILP), guarantee optimal solutions but hit a scalability wall at $n = 19$ due to the $O(n^3)$ constraint density and exponential search space. Conversely, standard model-free Reinforcement Learning (e.g., PPO) fails even earlier, finding optimal configurations only up to $10 \times 10$ grids before systematically failing at $n=11$~\citep{prellberg2025}. Generic RL agents struggle to maintain global geometric constraint awareness solely from scalar reward signals; a single collinear placement renders the entire configuration invalid, creating a sparse-reward ``validity cliff'' that frustrates gradient-based optimization. While \citet{prellberg2026constraintsatisfactionprogrammingnothreeinline} employs constraint satisfaction programming to search for configurations of size exactly $2n$, this incurs prohibitive computational costs, demanding 10 CPU days for $n \le 60$ and extensive GPU parallelization to merely reach $n=66$ \citep{flammenkamp_no3in_website}.

More recently, advanced AI paradigms like Large Language Models (FunSearch, \citealp{romera2023}; AlphaEvolve, \citealp{novikov2025alphaevolve}; and MCTS-AHD, \citealp{zheng2025montecarlotreesearch}) and Transformer-based search (PatternBoost, \citealp{charton2024patternboostconstructionsmathematicslittle}) have discovered impressive new constructions for similar extremal problems, for example the famous Cap Set Problem about points in $(\mathbb{Z}/3\mathbb{Z})^n$.  These models have not yet been successful when applied to the problems we consider here.  When PatternBoost is applied to study Max-N3IL, it matches optimal performance on small grids but fails to scale beyond $n=15$ \citep{prellberg2025}. This specific scalability wall arises because the transformer's pattern generalization is severely bottlenecked by the quality-speed trade-off inherent in the greedy algorithms used to generate its training data. Beyond this data generation bottleneck, transformer-based models are intrinsically highly \textit{token-consuming} and computationally expensive. Representing an $n \times n$ grid requires $O(n^2)$ tokens, leading to attention mechanisms that scale poorly for larger grids. Furthermore, autoregressive generation often struggles to strictly adhere to \textit{hard} global constraints without hallucinating invalid placements. LLM-driven methods like \textit{FunSearch} excel at finding concise heuristics, but it may be the case that optimal configurations on large grids are highly irregular and difficult to express using simple rules. Furthermore, potentially constrained by their training distributions, these LLMs may exhibit limitations in the out-of-distribution (OOD) algorithmic generalization required to discover genuinely novel logic for unsolved problems in this area  \citep{NEURIPS2024_4eff61b7, neuhaus2026outofdistributiongeneralizationreasoningmultimodal}. These limitations necessitate a different algorithmic paradigm that combines strict geometric constraint satisfaction with scalable search.

\paragraph{Why MCTS: A Geometry-Aware Approach}
To overcome the limitations of classical solvers and recent learning-based methods for problems like Max-N3IL, we propose a \textbf{Geometry-Aware MCTS} framework \citep{coulom2006efficient, kocsis2006bandit}. Monte Carlo Tree Search (MCTS) avoids the sparse-reward ``validity cliff'' of standard RL by using simulated trajectories to provide dense, constructive reward signals. Furthermore, it effortlessly enforces hard geometric constraints by masking invalid placements, bypassing both the hallucination risks and $O(n^2)$ token overhead of sequence models. While the asymmetric tree growth of MCTS naturally balances exploration and exploitation in unstructured spaces, scaling it to large grids (e.g., $n > 50$) requires specific adaptations. Our framework makes this tractable by (1) incrementally maintaining a strict feasible action space ($\mathcal{A}_\text{feas}$) to minimize per-node overhead, (2) exploiting $D_4$ grid symmetry during expansion and simulation to reduce branching, and (3) augmenting the search strategy with information persistence, adaptive exploration, and global optimum tracking to maximize yield under a finite computational budget.

\paragraph{Main Contributions}
Our work advances the intersection of combinatorial geometry and reinforcement learning by demonstrating the efficacy of search-based paradigms. By formulating extremal construction as a geometric Markov Decision Process (MDP), we introduce specific algorithmic innovations to make MCTS tractable in vast, constraint-heavy spaces. The specific contributions are as follows:

\begin{itemize}
    \item We propose an \textit{Incremental Feasible Action Space} ($\mathcal{A}_\text{feas}$) maintenance mechanism.  For problems based on constraints about collinear sets of points, such as Max-N3IL, this ray-casting approach reduces the per-node complexity for checking constraints from $O(n^3)$ to $O(n^2)$.  This effectively bypasses the sparse-reward ``validity cliff'' that typically paralyzes standard RL agents.

    \item We introduce \textit{Canonical Action Pruning} via dynamically computed state stabilizers $\text{Stab}(s) \leq D_4$, restricting tree expansion to orbit representatives while deliberately disabling pruning during rollouts to minimize computational overhead without sacrificing branching reduction.

    \item We propose \textit{Symmetric Batch Transitions}, a novel MDP transition mechanism that executes the simultaneous placement of an orbit under a chosen collection of symmetries. This significantly accelerates the discovery of promising symmetric configurations and allows canonical pruning to remain effective deeper into the search tree.

    \item We improve the best-known computational bounds for five out of the six problems that we considered.  Notably, we significantly extend the range in which we have lower bounds for Max-N3IL that are better than the asymptotic lower bound of size approximately $1.5n$.  We also find configurations giving upper bounds for the Smallest Complete Set problem that are smaller than the best-known construction.
\end{itemize}

\section{Problem Formulation}
\label{sec:problem_formulation}

We study extremal subset construction on the discrete grid $\mathcal{G}_n=[n]\times[n]$ under hard geometric constraints. Unlike in probabilistic generation tasks, a construction is either valid or invalid according to a binary predicate $\Phi$ (e.g., ``no three points are collinear''). We therefore cast the process as a sequential decision problem, enabling search and planning methods \cite{Vodopivec2017OnMC}.

\subsection{A General Geometric MDP Framework}
\label{subsec:mdp_framework}

We formulate geometric construction sequentially as a deterministic MDP defined by the tuple $(\mathcal{S}, \mathcal{A}, \mathcal{T}, \mathcal{R})$. Although the target configuration is permutation-invariant, this sequential formulation is computationally advantageous because it enforces hard constraints at every step, immediately pruning invalid branches. The state space $\mathcal{S} = \{s \mid s \subseteq \mathcal{G}_n\}$ is the power set of the grid, where a state $s \in \mathcal{S}$ represents a configuration of points, initialized at the empty set $s = \emptyset$. To maintain structural validity, we define a Boolean geometric invariant $\Phi: \mathcal{S} \to \{\text{True}, \text{False}\}$ (e.g., for Max-N3IL, $\Phi(s)$ holds iff no three points are collinear). Consequently, the action space is strictly confined to feasible placements, defined as $\mathcal{A}_\text{feas}(s) = \{p \in \mathcal{G}_n \setminus s \mid \Phi(s \cup \{p\}) = \text{True}\}$. This ensures that any action taken preserves $\Phi$ (efficient incremental maintenance of $\mathcal{A}_\text{feas}$ is detailed in Section~\ref{subsec:incremental_svas}). Transitions are deterministic and additive, given by $s' = s \cup \{a\}$ for $a \in \mathcal{A}_\text{feas}(s)$. A state $s$ is terminal (denoted as $s_T$) if and only if $\mathcal{A}_\text{feas}(s) = \emptyset$, meaning the configuration is maximal under inclusion.  Note that just because a configuration is maximal does not mean that it has optimal cardinality. Finally, the reward function $R(s_T, n)$ is evaluated at terminal states. For maximization variants (e.g., Max-N3IL), $R$ increases with $|s_T|$; for minimization variants (e.g., Min-Complete), it decreases with $|s_T|$. Detailed formulations are given in Appendix~\ref{subsec:reward_function}.

\subsection{Problem Variants}
\label{subsec:problem_variants}

We apply the proposed MDP framework to six extremal combinatorial geometry problems. For each problem, the geometric constraints are strictly governed by the state validity predicate $\Phi(s)$, and the objective is to either maximize or minimize the terminal set cardinality $|s_T|$. Let $P \subseteq s$ denote an arbitrary subset of the current state.

\begin{figure}[ht]
    \centering

    % --- First Subfigure ---
    \begin{subfigure}[t]{0.32\textwidth}
        \centering
        \includegraphics[width=0.55\linewidth]{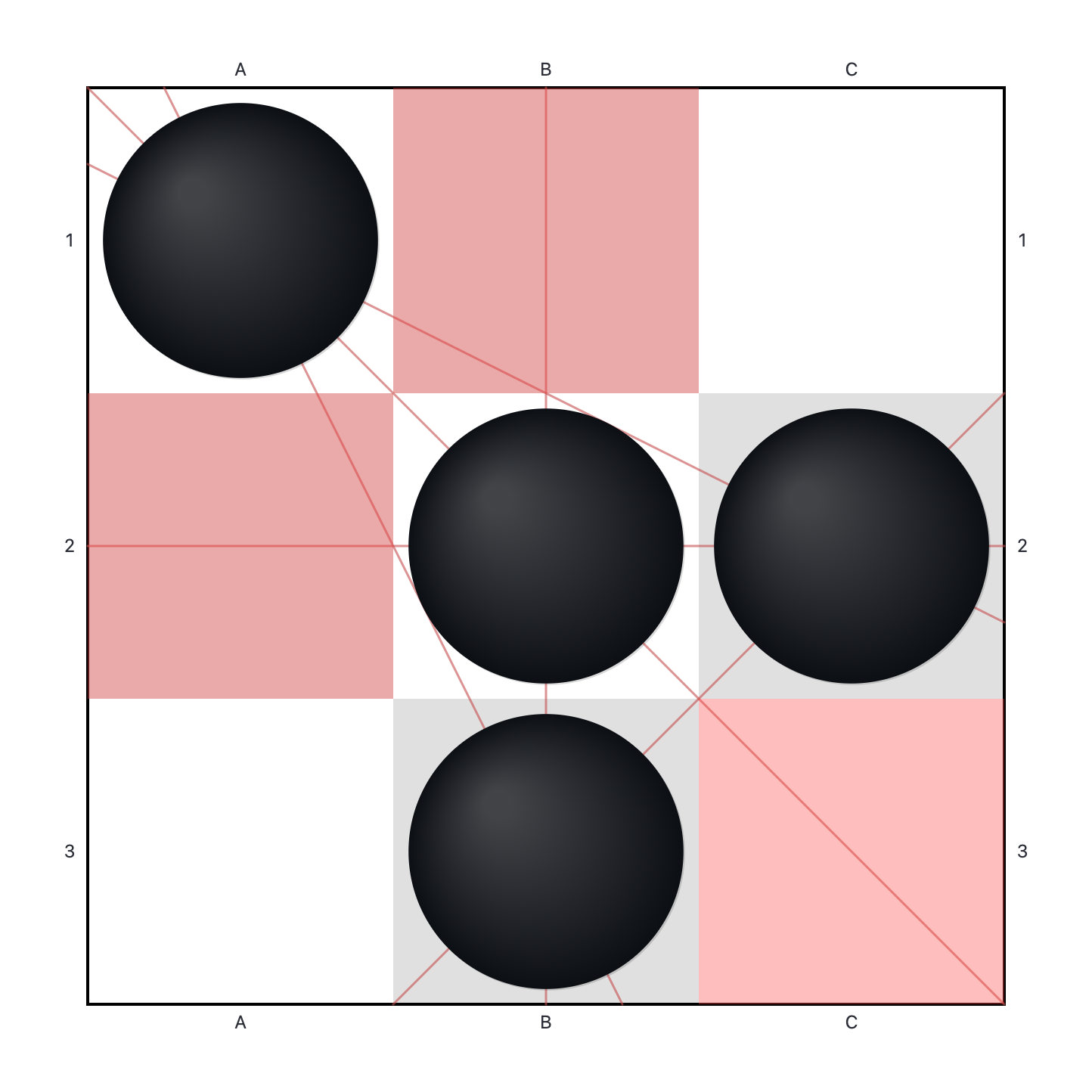}
        \caption{This set is not complete.  Either of the two white squares could be added without creating a collinear triple.}
        \label{fig:fig4}
    \end{subfigure}
    \hfill
    % --- Second Subfigure ---
    \begin{subfigure}[t]{0.32\textwidth}
        \centering
        \includegraphics[width=0.55\linewidth]{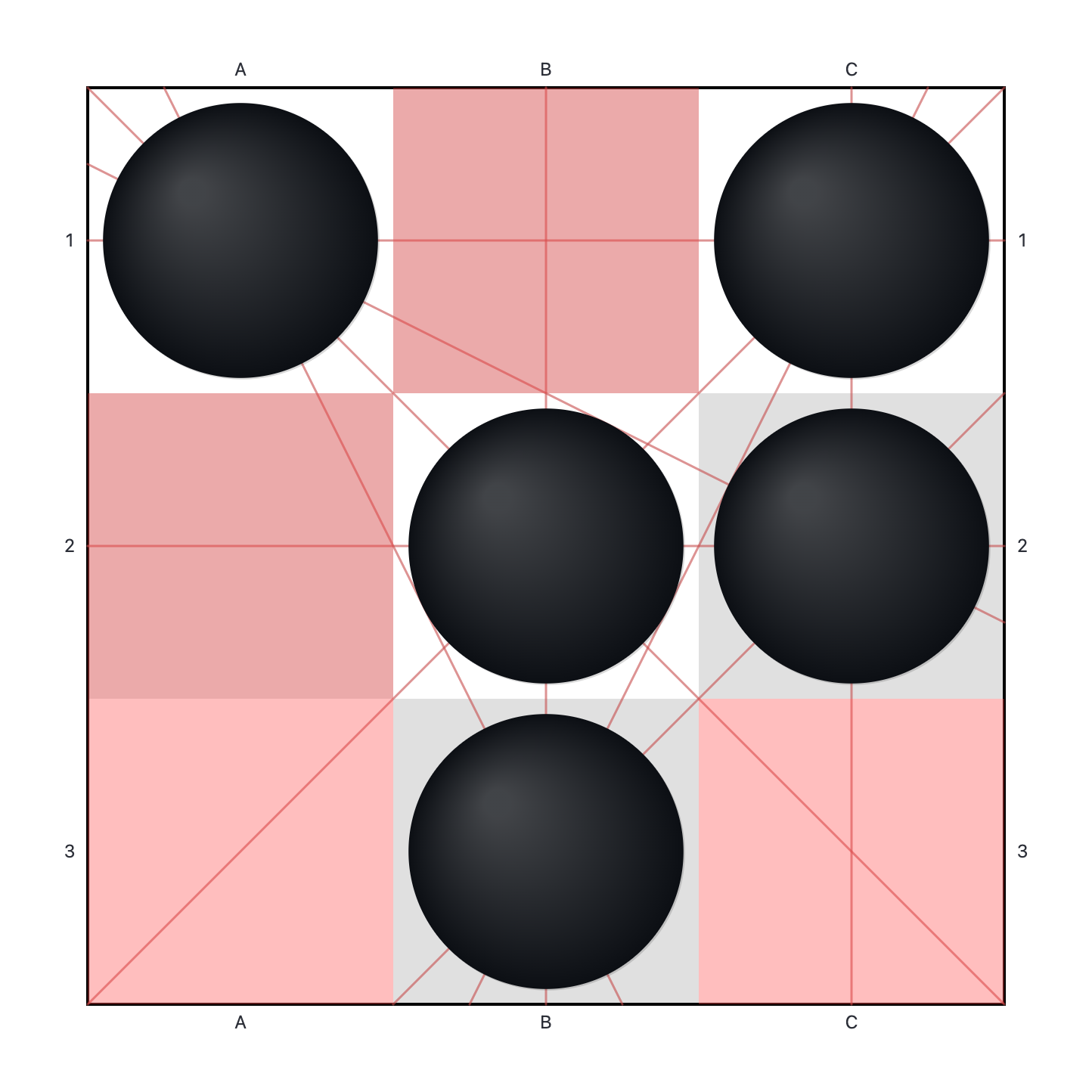}
        \caption{A complete set on a $3\times 3$ grid.}
        \label{fig:fig5}
    \end{subfigure}
    \hfill
    % --- Third Subfigure ---
    \begin{subfigure}[t]{0.32\textwidth}
        \centering
        \includegraphics[width=0.55\linewidth]{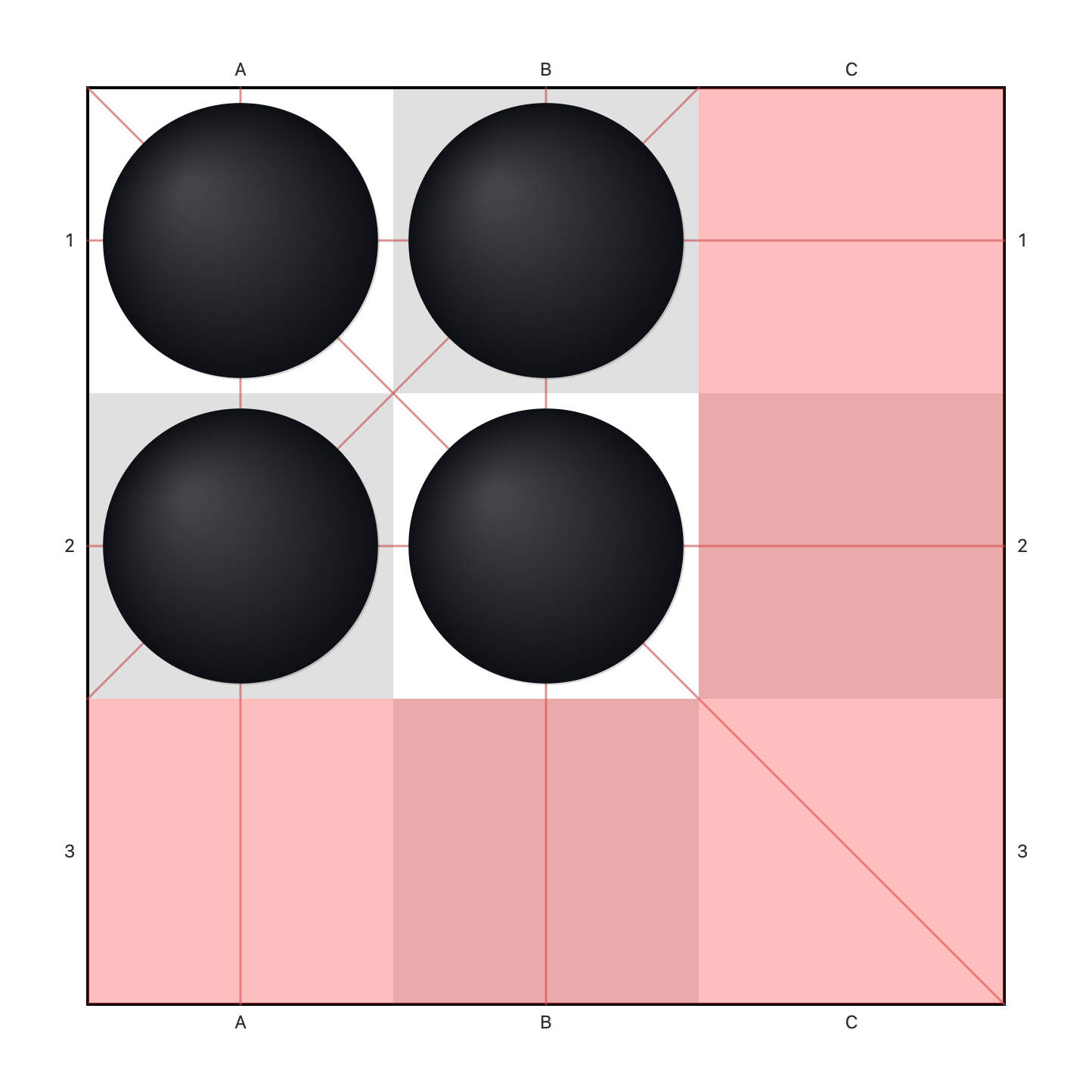}
        \caption{A complete set on a $3\times 3$ grid of minimum size.}
        \label{fig:fig6}
    \end{subfigure}

    \caption{Examples of incomplete and complete point configurations.}
    \label{fig:complete_sets}
\end{figure}

\paragraph{Collinear Constraints: Max-N3IL and Min-Complete.}
Both problems share the same predicate enforcing general position: $\Phi(s) = \left\{ \forall P \subseteq s \text{ with } |P|=3, \neg \text{Coll}(P) \right\}$, where $\text{Coll}(P)$ is true if the points in $P$ are collinear. Because our MDP naturally terminates only at inclusion-wise maximal states ($\mathcal{A}_\text{feas} = \emptyset$), any terminal configuration is inherently a ``complete'' set. As illustrated in \cref{fig:complete_sets}, the search stops exactly when all remaining empty cells lie on a line connecting two points already included in the set (\cref{fig:fig5,fig:fig6}). Max-N3IL seeks to maximize $|s_T|$ \citep{dudeney1917}, whereas Min-Complete (a.k.a. Smallest Independent Geometric Dominating Set) seeks to minimize $|s_T|$ \citep{AICHHOLZER2023101913}.

\paragraph{Smallest Geometric Dominating Set (Min-Dom).}
This problem is like Min-Complete, except that we drop the requirement that no three points of the configuration are collinear. Let $\text{Cover}(X)$ be true if the lines spanned by all pairs of distinct points in $X$ collectively cover every point in the grid $\mathcal{G}_n$ (i.e., every grid point is either in $X$ or collinear with at least two distinct points in $X$). The predicate is defined as $\Phi(s) = \left\{\forall s' \subsetneq s, \neg \text{Cover}(s')\right\}$. If $s$ is not a dominating set, then no subset of $s$ is a dominating set. If $s$ is a dominating set, then any superset of $s$ is also a dominating set.  In this way, we see that the search terminates exactly when a minimal dominating set is found. The objective is to minimize $|s_T|$ \citep{AICHHOLZER2023101913}.

\paragraph{Additional Problems.}
To demonstrate the flexibility of our framework, we explore three additional problems.  In each one, we aim to maximize $|s_T|$.
\begin{enumerate}
\item \textbf{Max-N4IL}: $\Phi(s) = \left\{ \forall P \subseteq s \text{ with } |P|=4, \neg \text{Coll}(P)\right\}$, a natural relaxation of N3IL.

\item \textbf{Max-No-4-on-Circle}: $\Phi(s) = \left\{\forall P \subseteq s \text{ with } |P|=4, \neg \text{Circ}(P)\right\}$, where $\text{Circ}(P)$ indicates the points are concyclic on a circle of finite radius \citep{dong2025largegridsubsetscospherical}.

\item \textbf{Max-No-Isosceles}: $\Phi(s) =\left\{ \forall P \subseteq s \text{ with } |P|=3, \neg \text{Isos}(P)\right\}$, where $\text{Isos}(P)$ is true if at least two Euclidean distances between points in $P$ are equal, including degenerate cases (i.e., collinear triples where one point is the midpoint of the segment).
\end{enumerate}

\section{Method: Geometry-Aware MCTS}
\label{sec:method}

Monte Carlo Tree Search (MCTS) has demonstrated remarkable versatility and success across diverse domains, ranging from significantly improving real-time decision-making in heuristic-free general video game playing \citep{Soemers_2016}, to enabling effective test-time scaling for Large Language Models \citep{wang2025mctsjudgetesttimescalingllmasajudge}. However, it is challenging to apply off-the-shelf MCTS to extremal problems in combinatorial geometry because of the strictness of the constraints and the $O(n^2)$ increase in the size of the action space. In this section, we first define the standard baseline and analyze its limitations, then introduce our Geometry-Aware framework.

\subsection{Baseline MCTS and the Naive Action Space}
\label{subsec:baseline}

We adopt the UCT (Upper Confidence bounds applied to Trees) algorithm \citep{kocsis2006bandit} as our baseline. During the selection phase, the agent traverses the tree by selecting actions that maximize the UCB1 (Upper Confidence Bound 1) criterion:
\begin{equation}
    \label{eq:ucb1}
    \text{UCB1}(s, a) = \frac{Q(s,a)}{N(s,a)} + C \sqrt{\frac{\ln N(s)}{N(s,a)}},
\end{equation}
where $Q(s,a)$ is the accumulated reward, $N(s,a)$ is the visit count, and $C$ governs exploration. Since our MDP transitions are deterministic, $Q(s,a)$ simplifies to the value of the unique successor state, $V(s'|s,a)$, and $N(s,a)$ precisely equals the visit count of that child node. After executing the standard MCTS phases (selection, expansion, simulation, and backpropagation) for a computational budget $H$, the search policy selects the most visited action: $a = \operatorname*{argmax}_{a \in \mathcal{A}(s)} N(s,a)$.

\paragraph{Naive Action Space and Dual Bottlenecks.}
Lacking domain-specific geometric constraints, the baseline agent considers any unoccupied grid cell as a potential move. This defines the \textbf{Naive Action Space}:
\begin{equation}
    \mathcal{A}_{\text{naive}}(s) = \mathcal{G}_n \setminus s.
\end{equation}
Operating on $\mathcal{A}_{\text{naive}}$ introduces two fundamental scalability bottlenecks. First, the \textbf{combinatorial branching factor} ($|\mathcal{A}_{\text{naive}}| \approx n^2$) severely dilutes visit counts at the root (e.g., nearly 5,000 actions for $n=70$), rendering UCB1 estimates high-variance. Second, the \textbf{validity bottleneck} arises because the density of valid moves decays rapidly. A rollout policy selecting uniformly from $\mathcal{A}_{\text{naive}}$ almost certainly violates the invariant $\Phi$, triggering immediate termination with sub-optimal reward. Consequently, discovering deep, valid configurations by chance becomes mathematically intractable for large $n$.

\subsection{Feasible Action Space via Incremental Maintenance}
\label{subsec:incremental_svas}

To resolve the validity bottleneck described in Section \ref{subsec:baseline}, our framework restricts the search strictly to the \textbf{Feasible Action Space} ($\mathcal{A}_{\text{feas}}$) that is described in Section \ref{subsec:mdp_framework}. This ensures that any action does not violate $\Phi$. Although $\mathcal{A}_{\text{feas}}(s)$ was defined conceptually in Section \ref{sec:problem_formulation}, computing it from scratch at every search node is computationally expensive. For the Max-N3IL problem, verifying a single candidate point $p$ against the current state $s$ can be done in $O(|s|) = O(n)$ time using slope hashing: one computes the normalized direction from $p$ to each $p_i \in s$ and checks for collisions in a hash table, where any collision witnesses a collinear triple. Aggregated over all $O(n^2)$ grid locations, a full validity scan from scratch therefore costs $O(n^3)$, which still becomes prohibitive when repeated at every node of a deep search tree.

\paragraph{Monotonicity and Incremental Updates.}
We leverage the property of \textit{constraint monotonicity}. In our target problems, placing a point never ``frees up'' a previously blocked location---it only imposes new constraints. Thus, $\mathcal{A}_{\text{feas}}(s') \subset \mathcal{A}_{\text{feas}}(s)$ for any successor state $s'$.
Instead of recomputing constraints from scratch, upon transitioning $s \to s \cup \{p_{new}\}$, we update $\mathcal{A}_{\text{feas}}$ by solely identifying cells invalidated by the new point:
\begin{equation}
    \mathcal{A}_{\text{feas}}(s \cup \{p_{new}\}) = \mathcal{A}_{\text{feas}}(s) \setminus \left( \{p_{new}\} \cup \text{NewConstraints}(p_{new}, s) \right).
\end{equation}

\begin{figure}[htbp]
    \centering
     % --- First Subfigure ---
    \begin{subfigure}[t]{0.32\textwidth}
        \centering
        \includegraphics[width=0.55\linewidth]{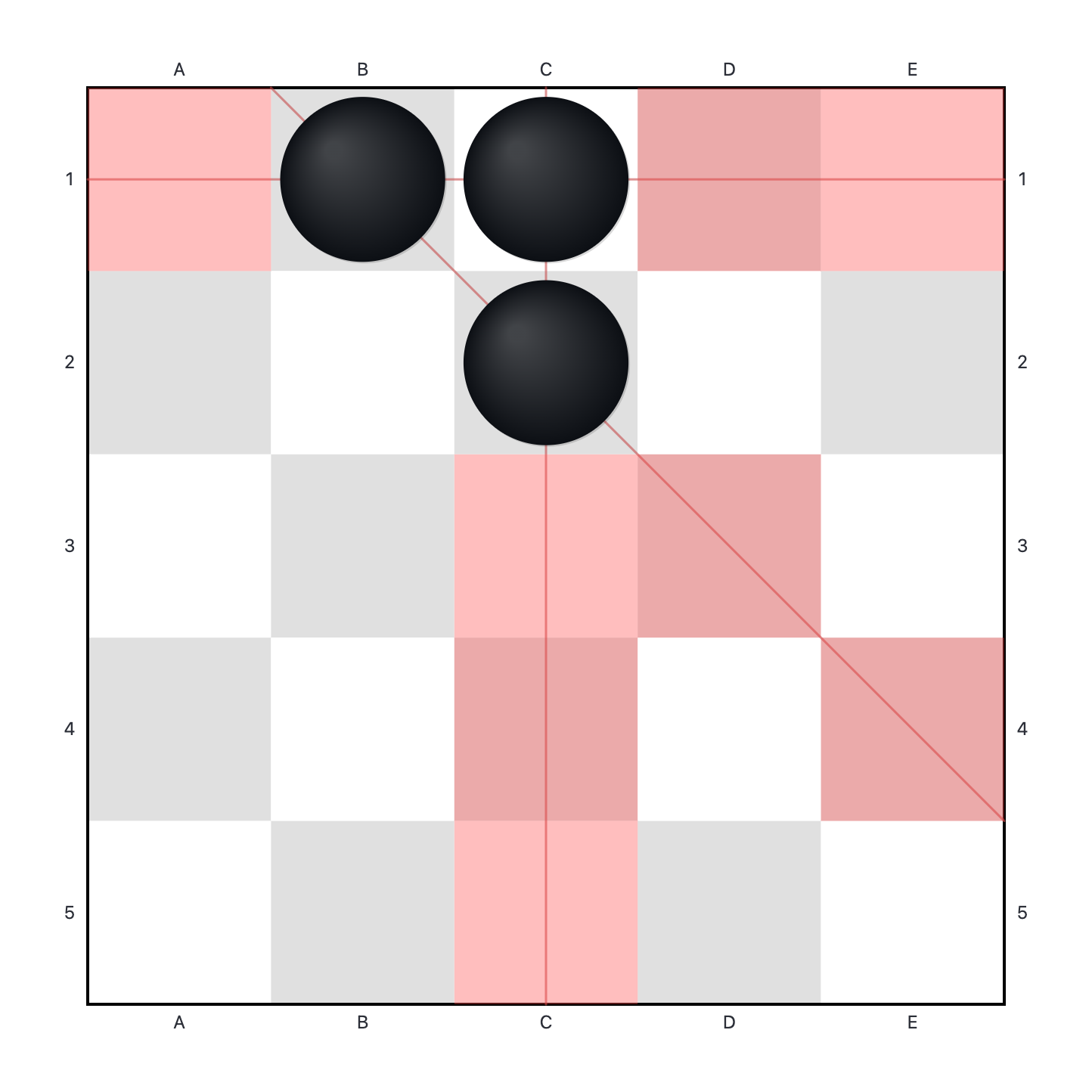}
        \caption{State $s$ with 3 points. Red cells indicate unoccupied points that are not in $\mathcal{A}_\text{feas}$.}
        \label{fig:feas_pts3}
    \end{subfigure}
    \hfill
    % --- Second Subfigure ---
    \begin{subfigure}[t]{0.32\textwidth}
        \centering
        \includegraphics[width=0.55\linewidth]{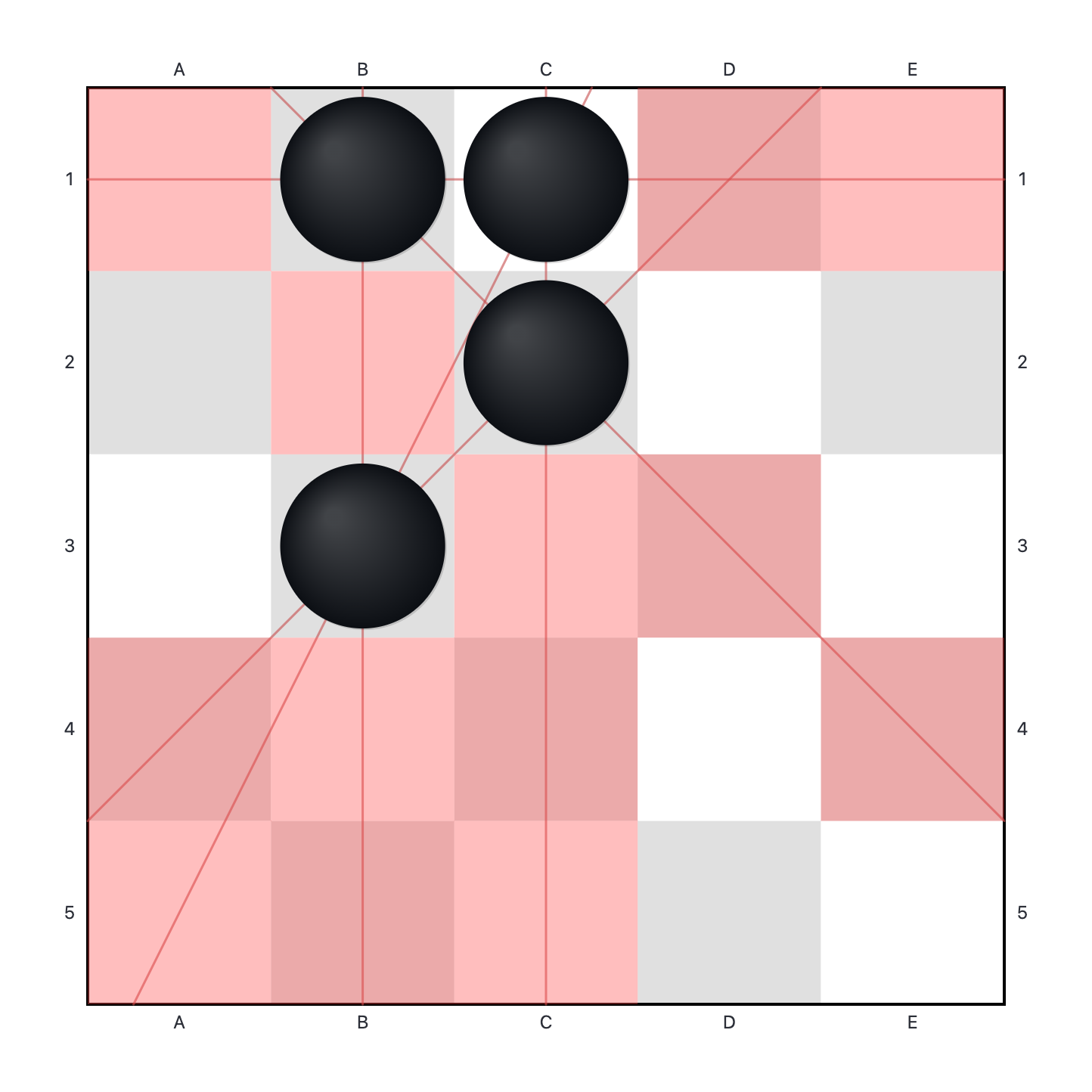}
        \caption{State $s'$ (4 points). Ray casting incrementally updates $\mathcal{A}_\text{feas}$.}
        \label{fig:feas_pts4}
    \end{subfigure}
    \hfill
    % --- Third Subfigure ---
    \begin{subfigure}[t]{0.32\textwidth}
        \centering
        \includegraphics[width=0.55\linewidth]{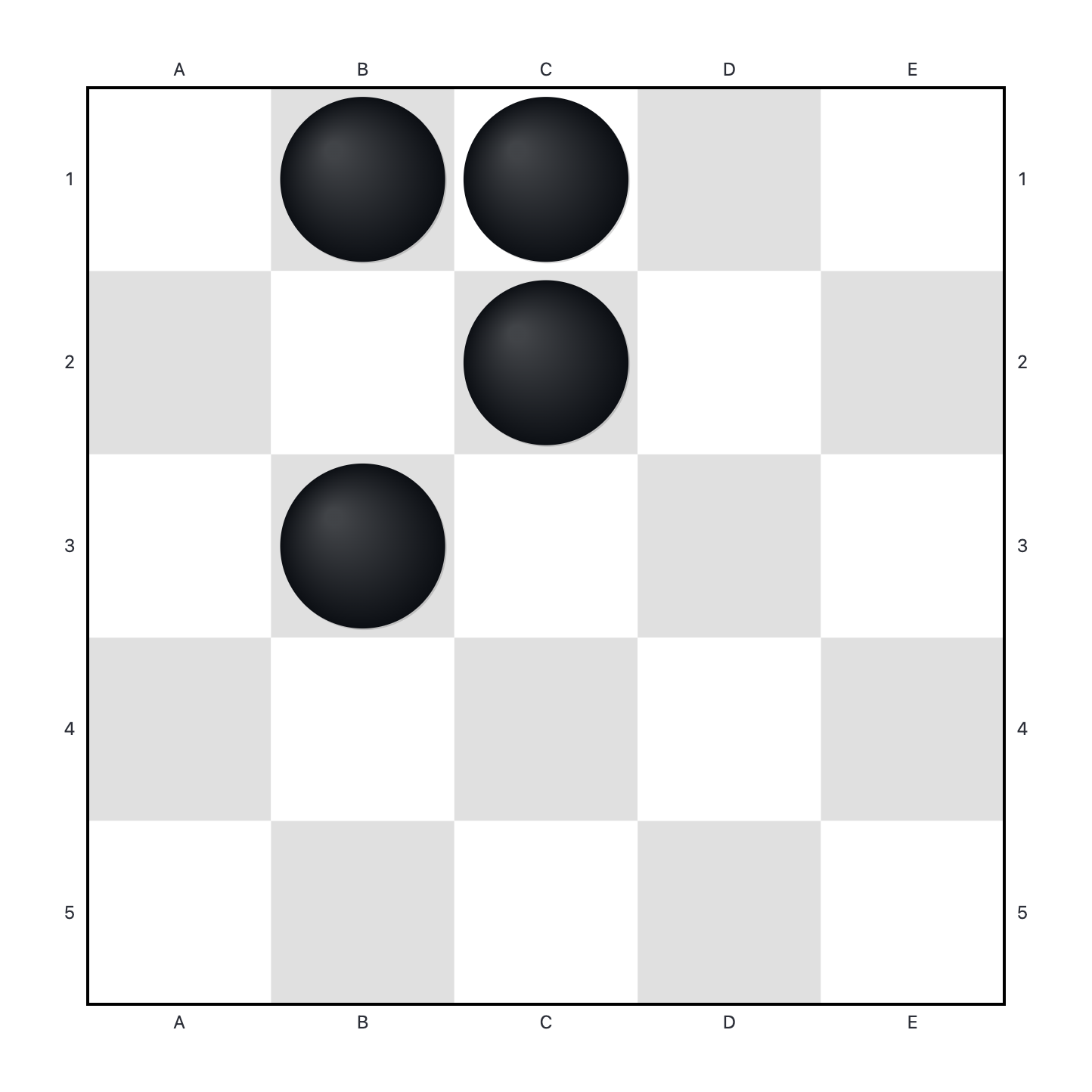}
        \caption{The naive action space $\mathcal{A}_\text{naive}$ for the same 4 points (only occupied cells are illegal).}
        \label{fig:naive_pts4}
    \end{subfigure}
    \caption{Visualizing the Incremental Ray Casting update and comparing $\mathcal{A}_\text{feas}$ with $\mathcal{A}_\text{naive}$ on a $5 \times 5$ grid.}
    \label{fig:incremental_update}
\end{figure}

\paragraph{Example: Ray Casting for Max-N3IL.}
For the Max-N3IL problem, this update takes the form of \textit{Ray Casting}. As illustrated in Figure \ref{fig:incremental_update}, placing a new point $p_{new}$ creates lines with each existing point $p_i \in s$. We write $\text{Ray}(u, v)$ for the set of points in the grid on the line connecting $u$ and $v$. The update rule becomes:
\begin{equation}
    \text{NewConstraints}(p_{new}, s) = \bigcup_{p_i \in s} \text{Ray}(p_{new}, p_i).
\end{equation}
Computing a discrete line (e.g., via GCD of coordinates) and marking its cells takes $O(n)$ operations, and there are $|s| \leq 2n$ existing points. Therefore, the total update complexity is $O(n^2)$. This represents a significant improvement over the $O(n^3)$ baseline, enabling deeper search within the same time budget.

Furthermore, the strict enforcement of geometric constraints naturally prunes the search tree. As illustrated by comparing Figure \ref{fig:feas_pts4} and Figure \ref{fig:naive_pts4}, the size of the feasible action space $|\mathcal{A}_\text{feas}|$ shrinks drastically as more points are placed, whereas $|\mathcal{A}_\text{naive}|$ only decreases by one per step. This effectively mitigates the \textbf{combinatorial branching factor} bottleneck identified in Section \ref{subsec:baseline}, allowing the UCB estimates to converge significantly faster on the remaining highly-promising branches.

\subsection{Exploiting Grid Symmetry}
\label{subsec:symmetry}

The $n \times n$ grid inherently possesses geometric symmetries described by the dihedral group $D_4$, which consists of eight isometric transformations (four rotations and four reflections). Because the validity predicate $\Phi$ is invariant under these transformations, the MDP induces highly redundant symmetric branches during tree search. We exploit this symmetry through two complementary mechanisms: reducing the search width via state-dependent canonical pruning, and accelerating the search depth via global symmetric batch transitions.

\subsubsection{Canonical Pruning via State Stabilizers}
\label{subsubsec:canonical_pruning}

To mitigate the combinatorial branching factor during the \emph{Expansion} phase of MCTS, we dynamically prune geometrically equivalent actions based on the symmetry of the current state. Let $\text{Stab}(s) \le D_4$ denote the stabilizer subgroup of a valid state $s$, defined as the set of transformations that leave the state configuration unchanged:
\begin{equation}
    \text{Stab}(s) = \{ g \in D_4 \mid g \cdot s = s \}.
\end{equation}
The group action of $\text{Stab}(s)$ partitions the feasible action space $\mathcal{A}_{\text{feas}}(s)$ into equivalence classes, or \emph{orbits}. For any candidate action $a \in \mathcal{A}_{\text{feas}}(s)$, its orbit under the stabilizer is denoted by $\mathcal{O}_{\text{Stab}(s)}(a) = \{g \cdot a \mid g \in \text{Stab}(s)\}$. Because transitioning to any action within the same orbit leads to isomorphic subtrees, we define a canonical pruning mechanism that restricts the expansion to exactly one representative action per orbit:
\begin{equation}
    \mathcal{A}_{\text{pruned}}(s) = \{\text{rep}(\mathcal{O}_{\text{Stab}(s)}(a)) \mid a \in \mathcal{A}_{\text{feas}}(s)\},
\end{equation}
where $\text{rep}(\cdot)$ is a deterministic tie-breaking function, such as selecting the action with the lexicographically smallest coordinate index.

Crucially, this canonical pruning is applied \emph{exclusively} during node expansion. We intentionally disable it during the \emph{Simulation} (rollout) phase. Since a simulation merely generates a single random trajectory to estimate the value function, computing stabilizers and orbits at each simulation step incurs severe computational overhead without the benefit of permanent branching factor reduction. The impact of this pruning is most impactful at the beginning of the search; for instance, at the empty root node where $|\text{Stab}(\emptyset)| = 8$, the initial branching factor is immediately reduced by nearly $7/8$ (as visualized in \cref{fig:sym_prun_empty_n3il}).

\subsubsection{Symmetric Batch Transitions}
\label{subsubsec:symmetric_batching}

While canonical pruning optimizes the tree width, empirical observations on the problems we tested suggest that optimal configurations frequently possess symmetries \citep{flammenkamp1992,flammenkamp1998,flammenkamp_no3in_website, prellberg2026constraintsatisfactionprogrammingnothreeinline,AICHHOLZER2023101913}. To incorporate this structural prior, we introduce \emph{Symmetric Batch Transitions}, which fundamentally redefines the MDP transition function to accelerate search depth and guide the agent toward highly symmetric subspaces.

\begin{figure}[htbp]
    \centering
     % --- First Subfigure ---
    \begin{subfigure}[t]{0.16\textwidth}
        \centering
        \includegraphics[width=\linewidth]{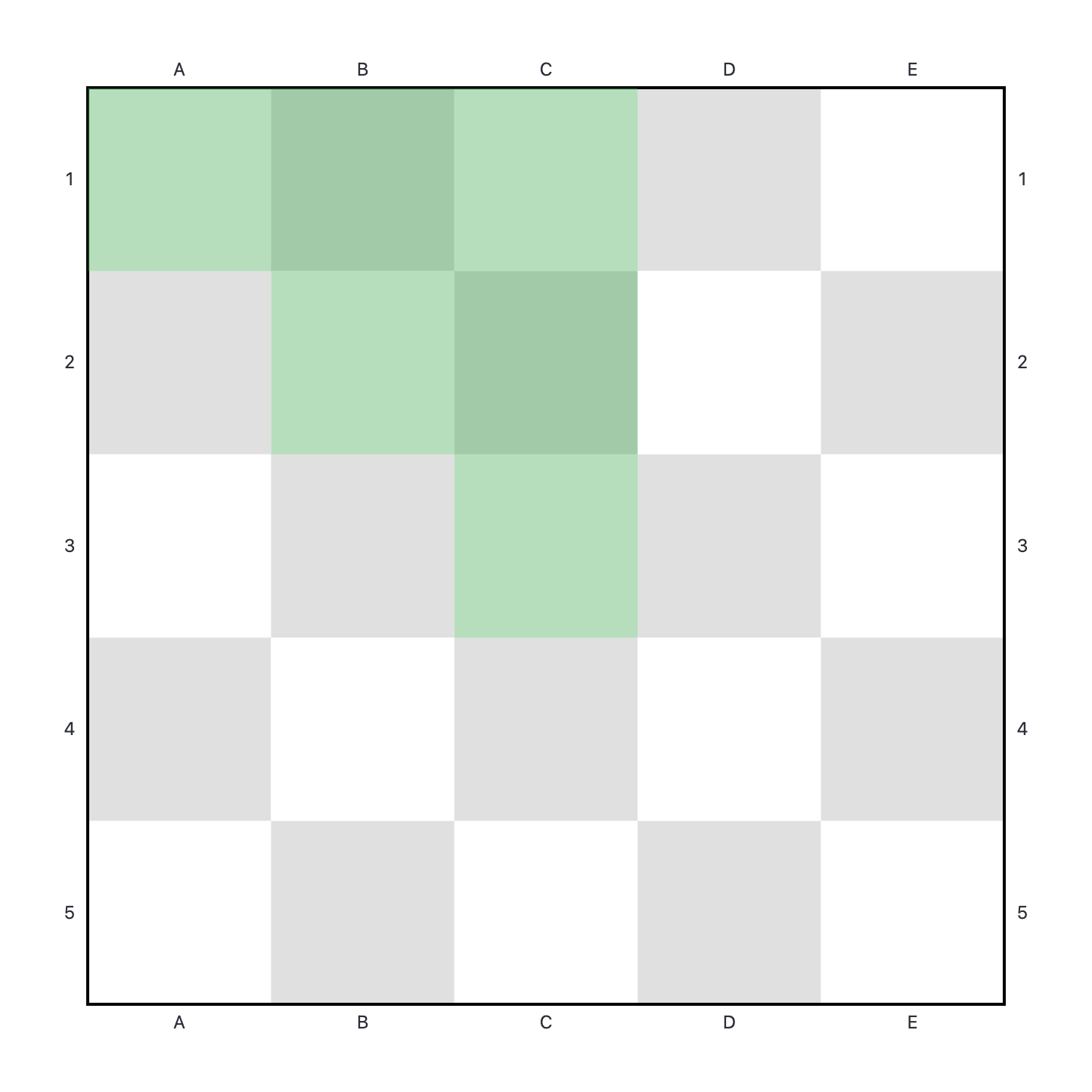}
        \caption{Pruned $\mathcal{A}_{\text{feas}}$ on an empty grid.}
        \label{fig:sym_prun_empty_n3il}
    \end{subfigure}
    \hfill
    % --- 2 Subfigure ---
    \begin{subfigure}[t]{0.16\textwidth}
        \centering
        \includegraphics[width=\linewidth]{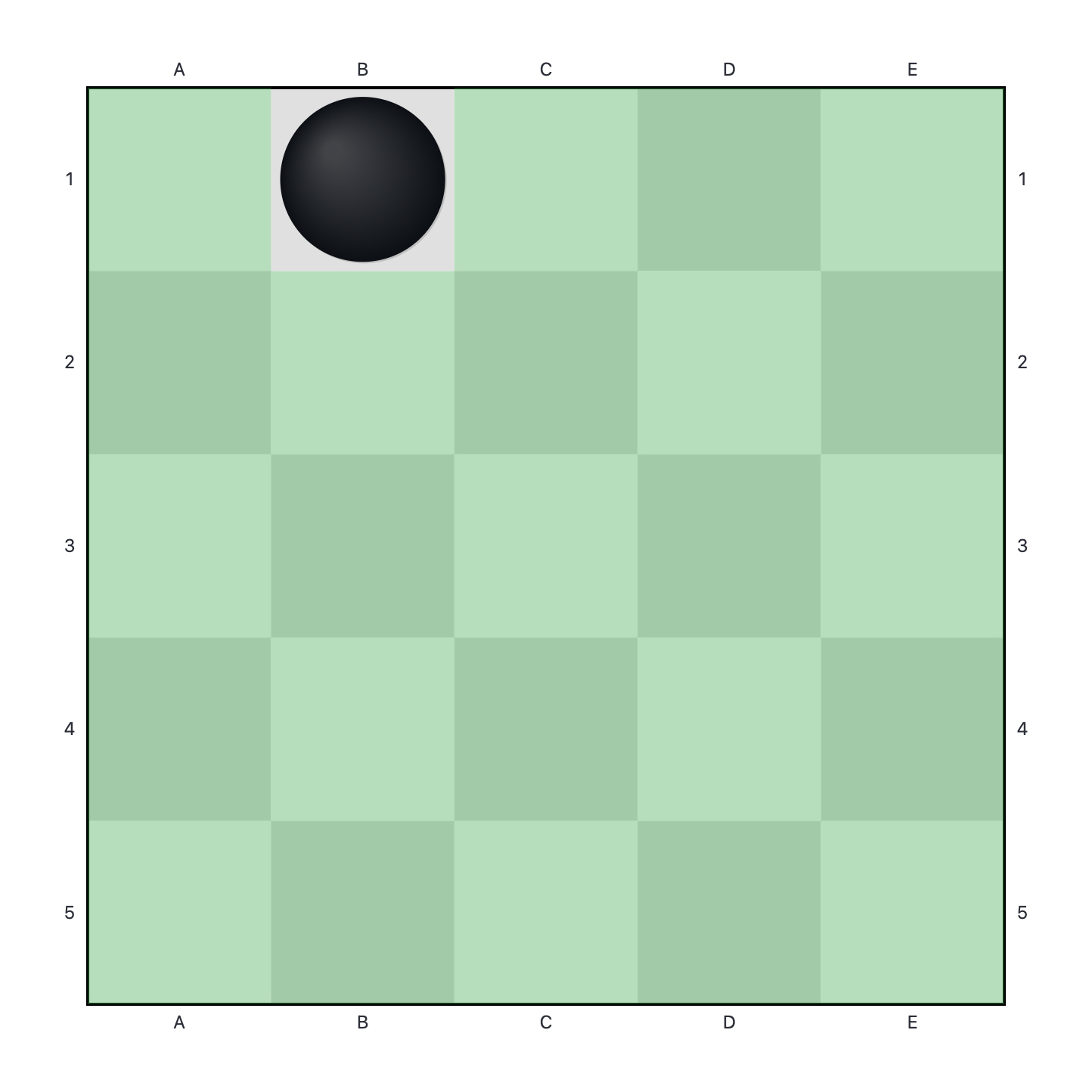}
        \caption{Place a point and attempt batch actions under $C_4$.}
        \label{fig:batch_act_1_n3il}
    \end{subfigure}
    \hfill
    % --- 3 Subfigure ---
    \begin{subfigure}[t]{0.16\textwidth}
        \centering
        \includegraphics[width=\linewidth]{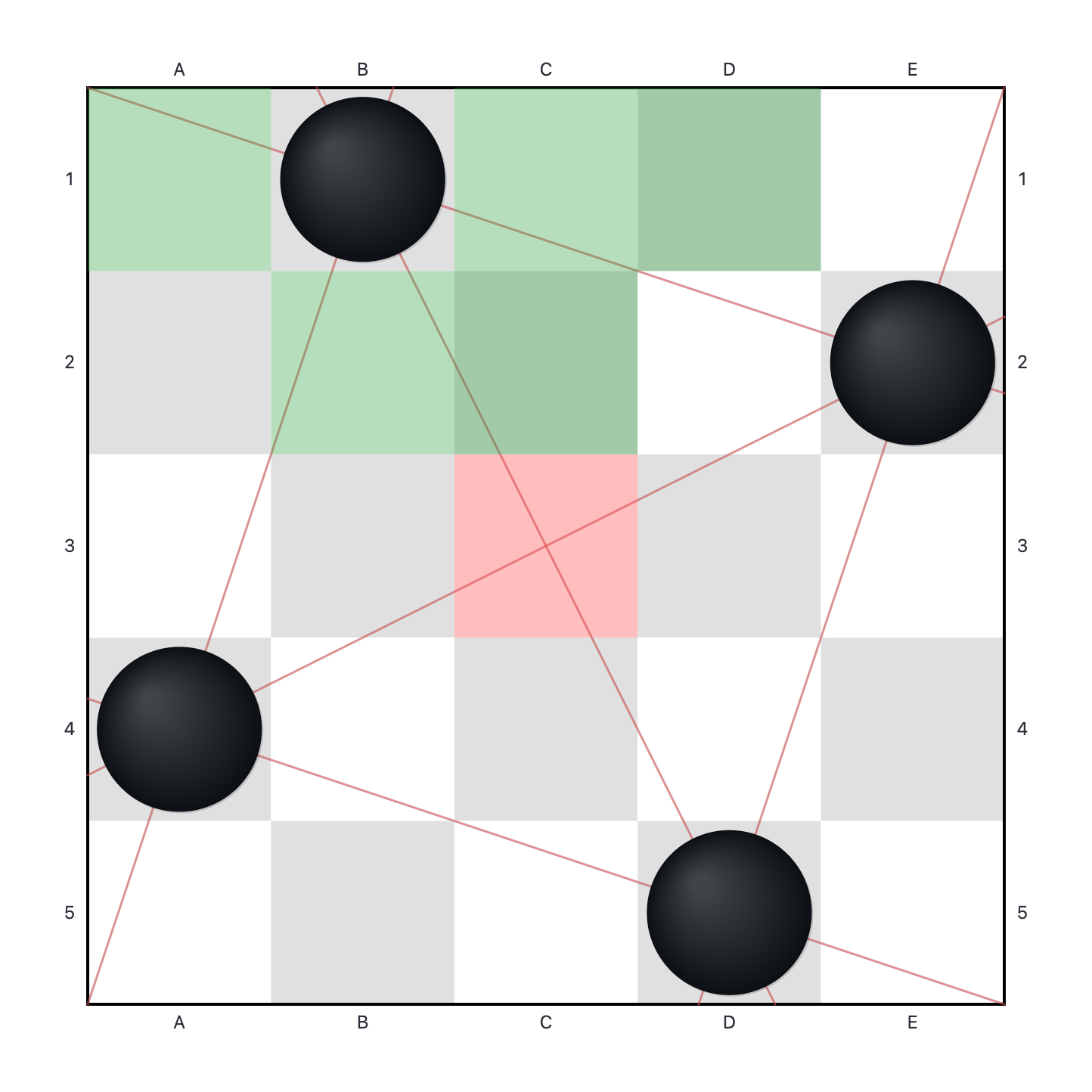}
        \caption{Batch transition successful.}
        \label{fig:sym_prun_4pts_n3il}
    \end{subfigure}
    \hfill
    % --- 4 Subfigure ---
    \begin{subfigure}[t]{0.16\textwidth}
        \centering
        \includegraphics[width=\linewidth]{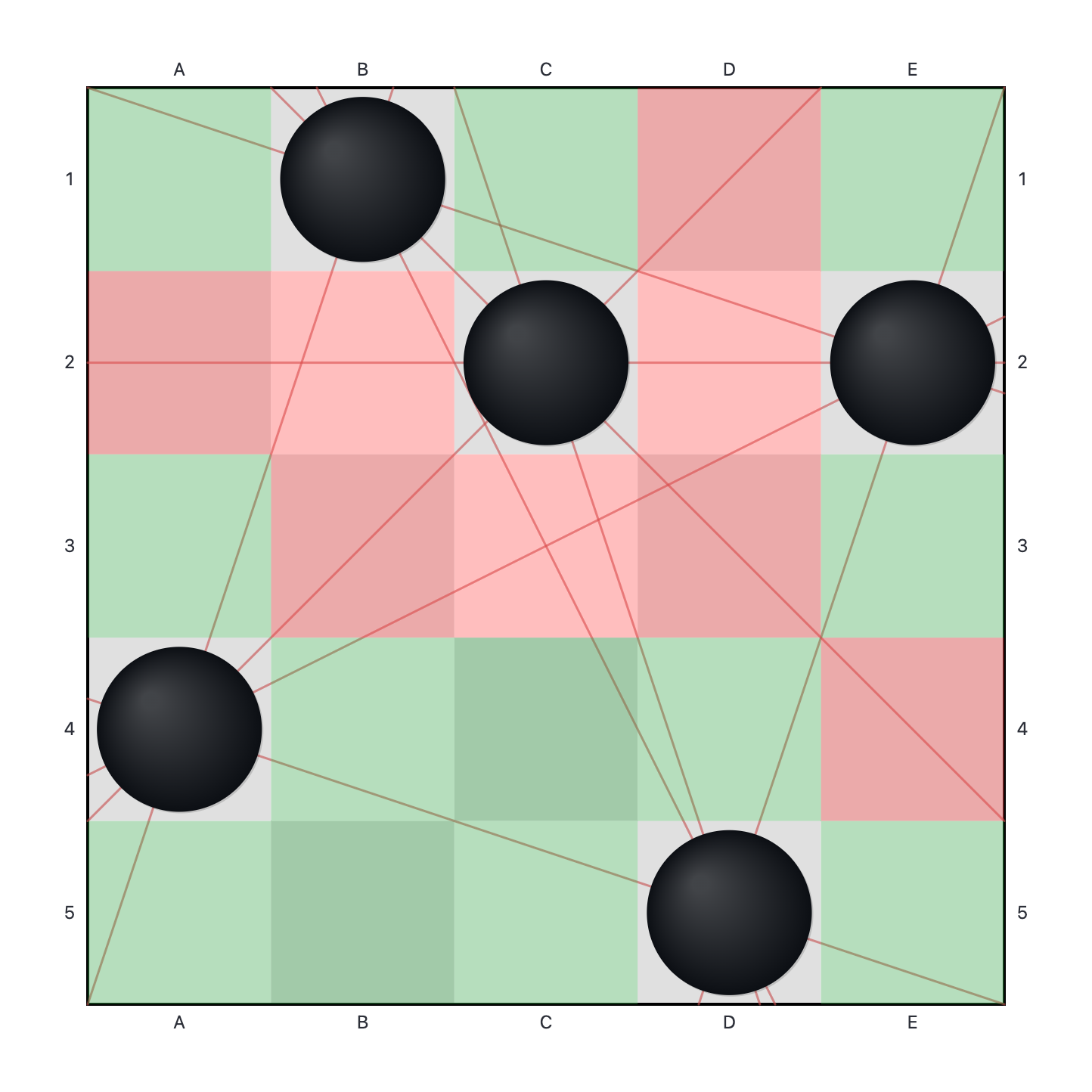}
        \caption{Place another point and try batch action under $C_4$.}
        \label{fig:batch_act2}
    \end{subfigure}
    % --- 5 Subfigure ---
    \begin{subfigure}[t]{0.16\textwidth}
        \centering
        \includegraphics[width=\linewidth]{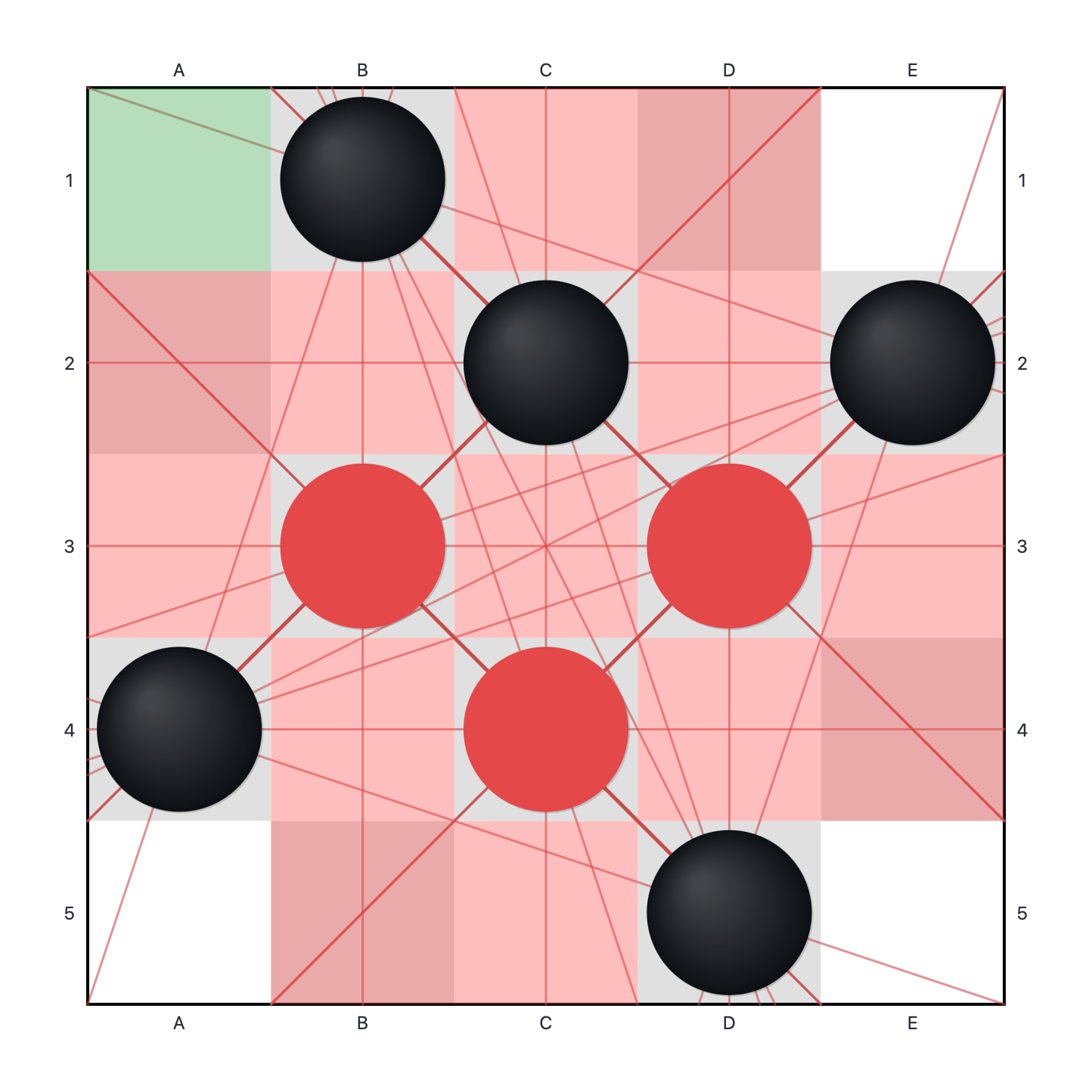}
        \caption{Constraint violated (red points created collinear triples).}
        \label{fig:batch_fail}
    \end{subfigure}
    % --- 6 Subfigure ---
    \begin{subfigure}[t]{0.16\textwidth}
        \centering
        \includegraphics[width=\linewidth]{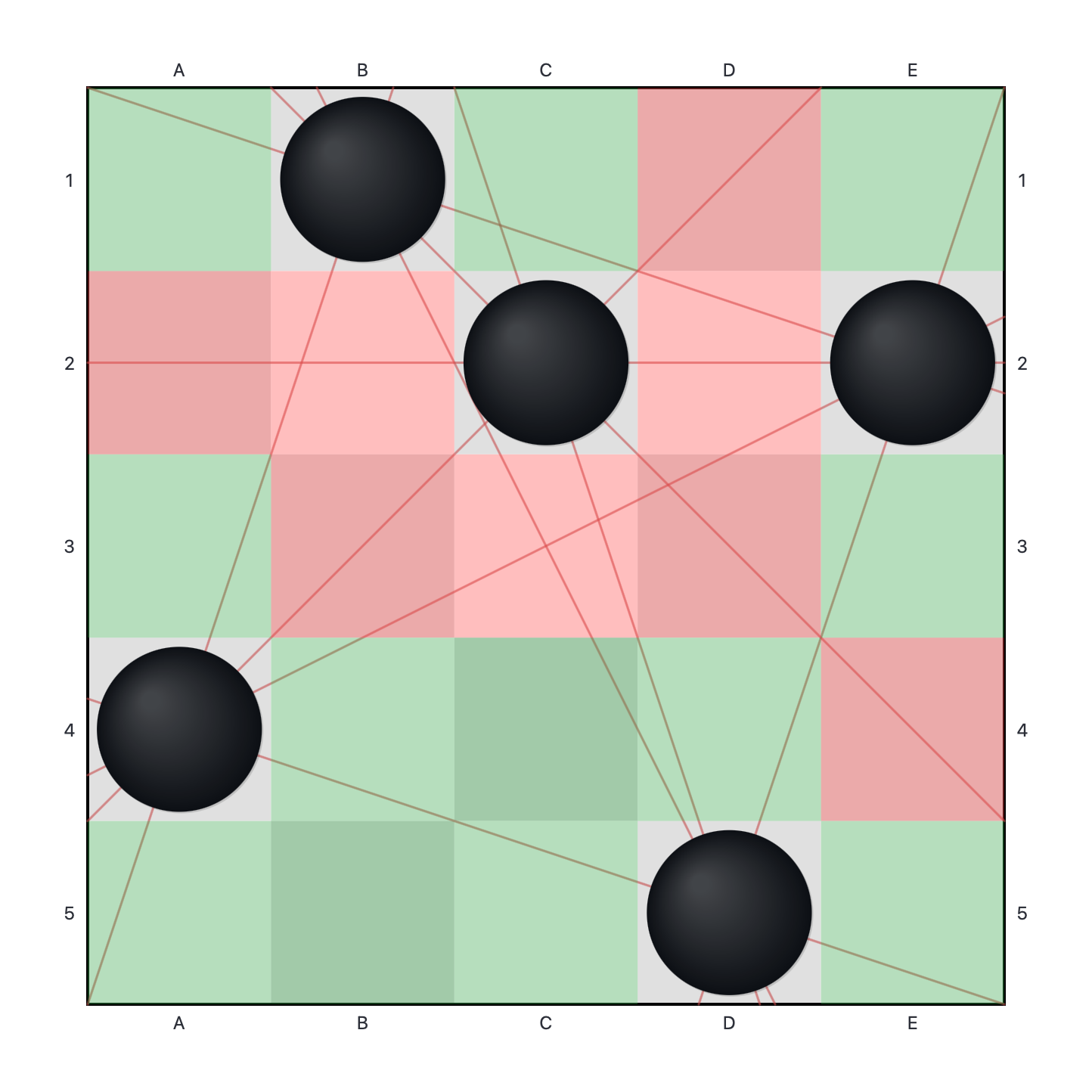}
        \caption{Fallback to single action.}
        \label{fig:batch_fallback}
    \end{subfigure}

    \caption{Visualization of Canonical Pruning via State Stabilizers (green-shaded spaces are pruned $\mathcal{A}_{\text{feas}}$) and Symmetric Batch Transitions (succeeded from action \subref{fig:batch_act_1_n3il} to state \subref{fig:sym_prun_4pts_n3il} but failed from action \subref{fig:batch_act2} to state \subref{fig:batch_fail}). Note that canonical pruning is effective on states \subref{fig:sym_prun_empty_n3il} and \subref{fig:sym_prun_4pts_n3il}.}
    \label{fig:sym_pruning_and_batch_transition}
\end{figure}

Before the search begins, we pre-select a group of symmetries $G \le D_4$ (e.g., the rotational group $C_4$). When an action $a \in \mathcal{A}_{\text{feas}}$ is selected by the agent (either in the tree or during rollout), we compute its orbit under the action of $G$, denoted as $\mathcal{O}_G(a) = \{g \cdot a \mid g \in G\}$. We then evaluate whether these points can be placed simultaneously without violating the geometric invariant. The modified transition function $\mathcal{T_\text{batch}}$ follows an ``all-or-nothing'' rule:
\begin{equation}
    \mathcal{T}_{\text{batch}}(s) =
    \begin{cases}
        s \cup \mathcal{O}_G(a) & \text{if } \Phi(s \cup \mathcal{O}_G(a)) = \text{True}, \\
        s \cup \{a\} & \text{otherwise}.
    \end{cases}
\end{equation}
The batch transition is executed internally as a sequence of incremental additions, ensuring that each point is validated against the updated feasible action space $\mathcal{A}_{\text{feas}}$. As a result, the agent commits the entire orbit $\mathcal{O}_G(a)$ in a single MDP transition if mutually compatible (\cref{fig:batch_act_1_n3il,fig:sym_prun_4pts_n3il}); otherwise, it gracefully falls back to placing only the single action $a$ (\cref{fig:batch_act2,fig:batch_fail,fig:batch_fallback}).

This batching mechanism operates throughout the entire MCTS pipeline, drastically accelerating terminal state evaluations during simulations. Furthermore, it creates a powerful synergy with the canonical pruning described in Section \ref{subsubsec:canonical_pruning}. Batch transitions are designed to generate states that preserve the global symmetry $G$ deeper within the search tree, provided the fallback is not triggered. Consequently, canonical pruning remains highly effective at reducing $|\mathcal{A}_{\text{feas}}|$ at deeper search depths.

\subsection{Enhanced Search Strategy}
\label{subsec:enhanced_search}

To further maximize the performance of Geometry-Aware MCTS under a finite computational budget, we adopt and adapt three targeted optimization strategies. These enhancements focus on \emph{information persistence}, \emph{dynamic exploration adjustment}, and \emph{global result harvesting}, transforming the standard MCTS into a more efficient solver for deterministic combinatorial tasks.

\subsubsection{Subtree Reuse (Information Persistence)}
\label{subsubsec:tree_reuse}

In standard MCTS implementations, the search tree is typically discarded after each decision step, resetting the agent's knowledge to zero. Given that our MDP transitions are deterministic, discarding the tree wastes significant computational effort. We therefore adopt a \textbf{Subtree Reuse} strategy \citep{silver2016mastering}. Upon executing an action $a$ and transitioning from the current root $s$ to a child state $s'$, we do not re-initialize the tree. Instead, we promote the child node corresponding to $s'$ to become the new root for the subsequent search phase. To maintain memory efficiency, the previous root and its unselected branches are detached and garbage-collected. This strategy ensures that the new search phase begins with a ``warm start,'' where $N_{\text{existing}}(s') > 0$, allowing the agent to immediately focus on deeper lookaheads rather than re-exploring the immediate action space.

\subsubsection{Time-Dependent Exploration Decay}
\label{subsubsec:exploration_decay}

A critical challenge in the extremal problems we study is the \emph{small suboptimality gap}: the value difference between optimal and near-optimal actions is often extremely small. Even with exponential reward shaping, the exploration term $C \sqrt{\ln N(s) / N(s,a)}$ in the UCB1 formula (\ref{eq:ucb1}) can dominate the exploitation term $Q(s,a)/N(s,a)$, hindering convergence. To address this, we implement a time-dependent exploration decay mechanism \citep{ruijl2013combiningsimulatedannealingmonte}, modulating the exploration constant via a decay function $\lambda(r)$:
$
    C(t) = C_{\text{base}} \cdot \lambda(r),\text{ where } \lambda(r) = 1 - 0.85\sqrt{r}.
$
Here, $r = i/N_{\text{total}} \in [0, 1]$ represents the progress ratio of the search, where $i$ is the current iteration index and $N_{\text{total}}$ is the allocated search budget. A critical design choice is formulating $r$ under the Subtree Reuse paradigm. If $r$ is computed relative only to the newly allocated budget for the current step, $C(t)$ resets to its maximum at every transition, destroying accumulated confidence. Instead, we incorporate the inherited visit count $N_{\text{existing}}(s')$ from the previous tree as valid prior exploration:
\begin{equation}
    r = \frac{N_{\text{existing}}(s') + i}{N_{\text{total}}}.
\end{equation}
This formulation ensures the exploration factor $C(t)$ remains continuous and monotonically decreasing across sequential decisions. As the effective visit count approaches $N_{\text{total}}$, the decay aggressively reduces exploration noise, committing the policy to the empirically superior branch.

\subsubsection{Anytime Optimal State Tracking}
\label{subsubsec:optimal_tracking}

The standard output of MCTS is a policy derived from the visit counts of the root node, intended to maximize the expected future reward. However, in extremal problems, our objective is to identify the \emph{global maximum} configuration (e.g., the largest possible set), regardless of whether it lies on the most probable path. To address this, we augment MCTS with an \textbf{Anytime Optimal State Tracking} mechanism. During the simulation phase, whenever the rollout policy reaches a terminal state $s_T$, we evaluate its reward $R(s_T,n)$. If $R(s_T,n)$ exceeds the best reward found so far globally, we record the state $s_T$, independent of the tree search statistics. This decoupling ensures that valid ``outlier'' solutions found via serendipity in deep rollouts are never lost, effectively treating the simulation phase as a global random sampler running in parallel with the policy optimization.

\section{Experiments and Results}

\label{sec:results}

\subsection{Experimental Setup}

\label{subsec:exp_setup}

We conducted all experiments on a High Performance Computing Cluster using a single CPU core and a strict 6GB memory limit per trial.  We chose this limit to show the memory efficiency of our framework. Time limits are set to 14 days for Max-N3IL and 7 days for the other problems we tested. Comprehensive hardware details, hyperparameter configurations, and ablation constraints are provided in Appendix \ref{appendix:detailed_exp_configurations}~and~\ref{appendix:tech_details}.

Our baseline strategy was a single search trial per configuration to maximize the exploration of large grid sizes under our strict compute budget. However, due to opportunistic job scheduling on the shared cluster, a subset of configurations received multiple evaluations. Because our objective is to establish theoretical existence bounds rather than expected average returns, we report the absolute best (global maximum or minimum) valid configuration discovered across all available trials for any given $n$.

\subsection{Main Results}

\begin{table}[ht]
\centering
\caption{Comparison of our MCTS results against best-known theoretical and computational bounds. See Appendix \ref{sec:related_works} and \ref{appendix:detailed_previous_best_known} for detailed previous best known results. \textbf{Note:} ``Ours'' reports the empirical performance trend observed or specific improvements found within the tested range. All baseline comparisons are based on the records available as of June 23, 2026.}
\label{tab:sota_comparison}
\resizebox{\textwidth}{!}{%
\begin{tabular}{l l l l l}
\toprule
\textbf{Extremal Problem} & \textbf{Metric} & \textbf{Previous Best Known (Baseline)} & \textbf{Ours} & \textbf{Improved Range ($n$)} \\
\midrule
\multirow{2}{*}{\textbf{Max No-Three-in-Line}} & \multirow{2}{*}{Lower Bound} & $(1.5-\varepsilon)n$ \citep{hall1975} & \multirow{2}{*}{$\mathbf{\approx 1.8n}$ (Empirical Fit)} & \multirow{2}{*}{$\ourNothreeILnewbestknownbeginN \le n \le 119$} \\
& & See \cref{tab:best_known_n3il_detailed} for known optimal and near-optimal results &  & \\
\midrule
\textbf{Min Complete} & Upper Bound & See Table \ref{tab:optimal_smallest_complete_2_to_12} and \ref{tab:current_smallest_complete_13_and_more} & $\approx \mathbf{0.95n}$ (Empirical Fit) & $37 \leq n \leq 96$ \\
\midrule
\multirow{2}{*}{\textbf{Min Dominating Set}} & \multirow{2}{*}{Upper Bound} & $2\lceil n/2 \rceil$ \citep{AICHHOLZER2023101913} & \multirow{2}{*}{$\approx \mathbf{0.95n}$ (Empirical Fit)} & $37 \leq n \leq 96,$ $31 \leq n \leq 35,$ \\
& & For small grids, see Table \ref{tab:optimal_smallest_complete_2_to_12} and \ref{tab:current_smallest_complete_13_and_more} &  & and $n\in\{7, 16, 26$\} \\
\midrule
\textbf{Max No-Four-in-Line} & Lower Bound & $1.973n$ for sufficiently large $n$ \citep{kovacs2025randomised} & $\mathbf{= 3n}$ & $ n \le 100$ \\
\midrule
\multirow{2}{*}{\textbf{Max No Isosceles Triangle}} & \multirow{2}{*}{Lower Bound} & $\Omega(n/\sqrt{\log n})$ \citep{charton2024patternboostconstructionsmathematicslittle} & \multirow{2}{*}{$\approx 1.4n$ (Empirical Fit)} & - \\
 & & $112$ for $n=64$ and $164$ for $n=100$ \citep{georgiev2025mathematicalexplorationdiscoveryscale} &  &  \\
\midrule
\textbf{Max No-Four-on-a-Circle} & Lower Bound & $n-o(n)$ \citep{dong2025largegridsubsetscospherical} & $\approx \mathbf{1.8n}$ (Empirical Fit)  & $n\leq 39$ \\
\bottomrule
\end{tabular}%
}
\end{table}

\begin{figure}[ht]
    \centering

    % --- Row 1 ---
    \begin{subfigure}[t]{0.32\textwidth}
        \centering
        \includegraphics[width=\linewidth]{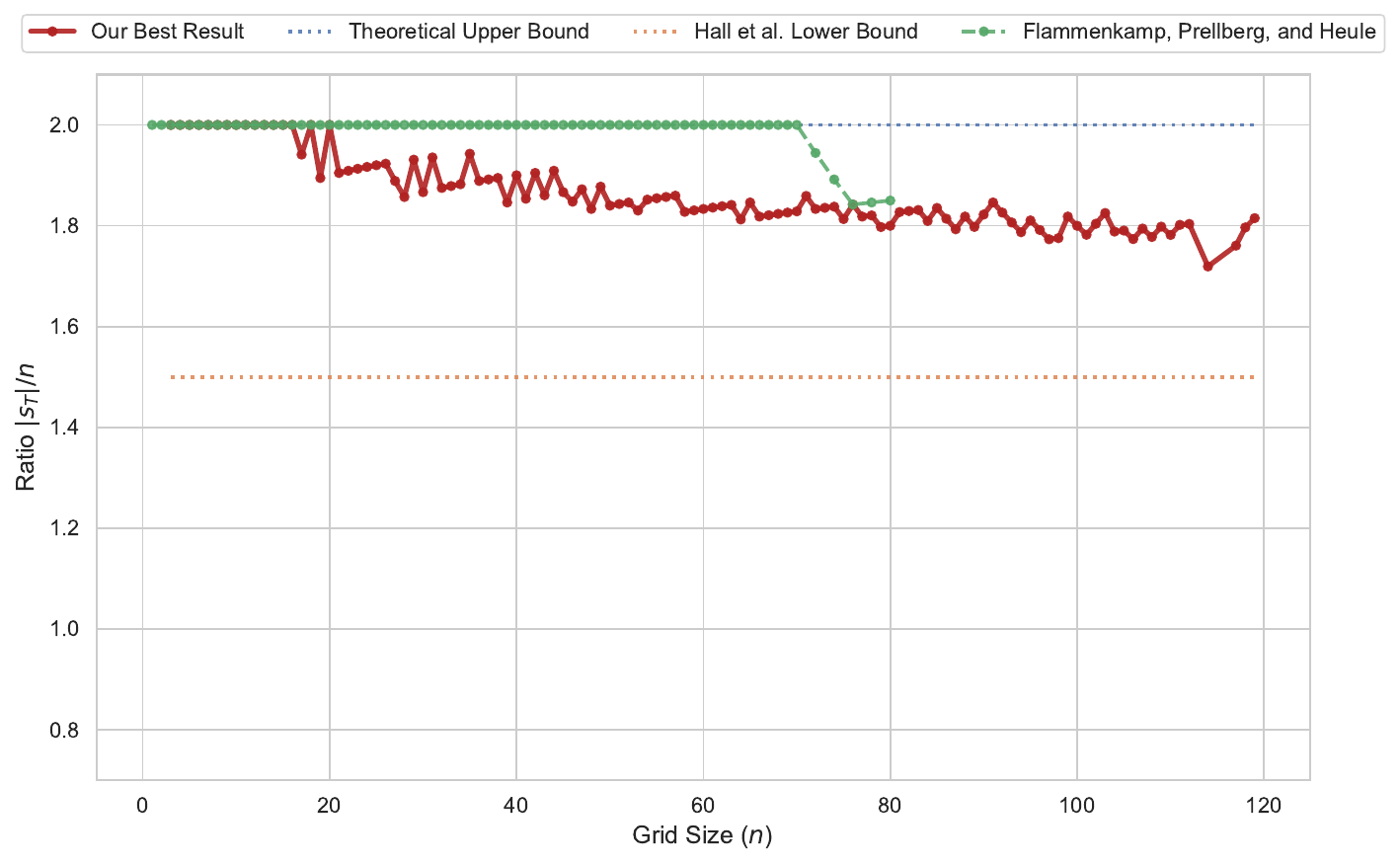}
        \caption{Max-N3IL}
        \label{fig:res_max_n3il}
    \end{subfigure}
    \hfill
    \begin{subfigure}[t]{0.32\textwidth}
        \centering
        \includegraphics[width=\linewidth]{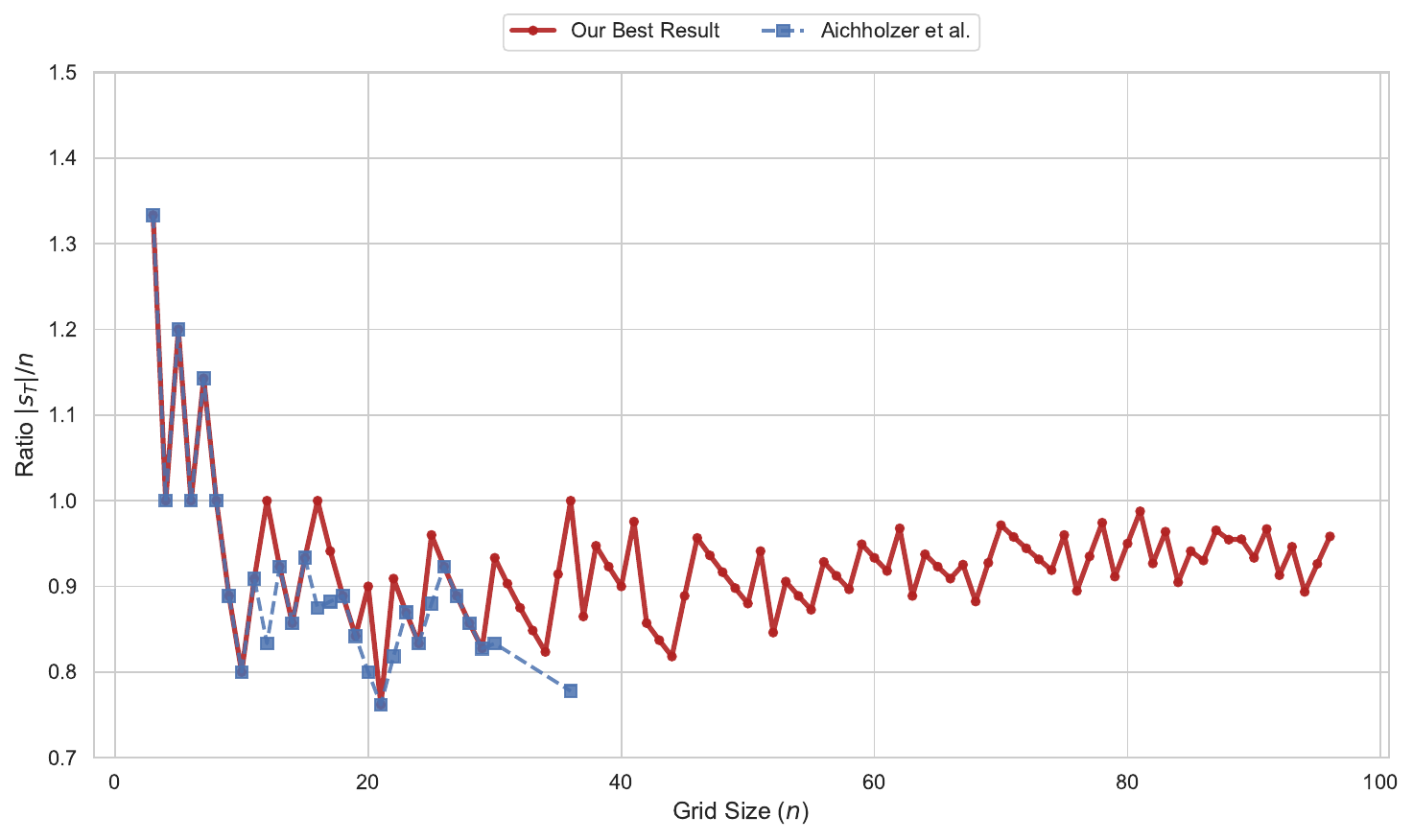}
        \caption{Min-Complete}
        \label{fig:res_min_complete}
    \end{subfigure}
    \hfill
    \begin{subfigure}[t]{0.32\textwidth}
        \centering
        \includegraphics[width=\linewidth]{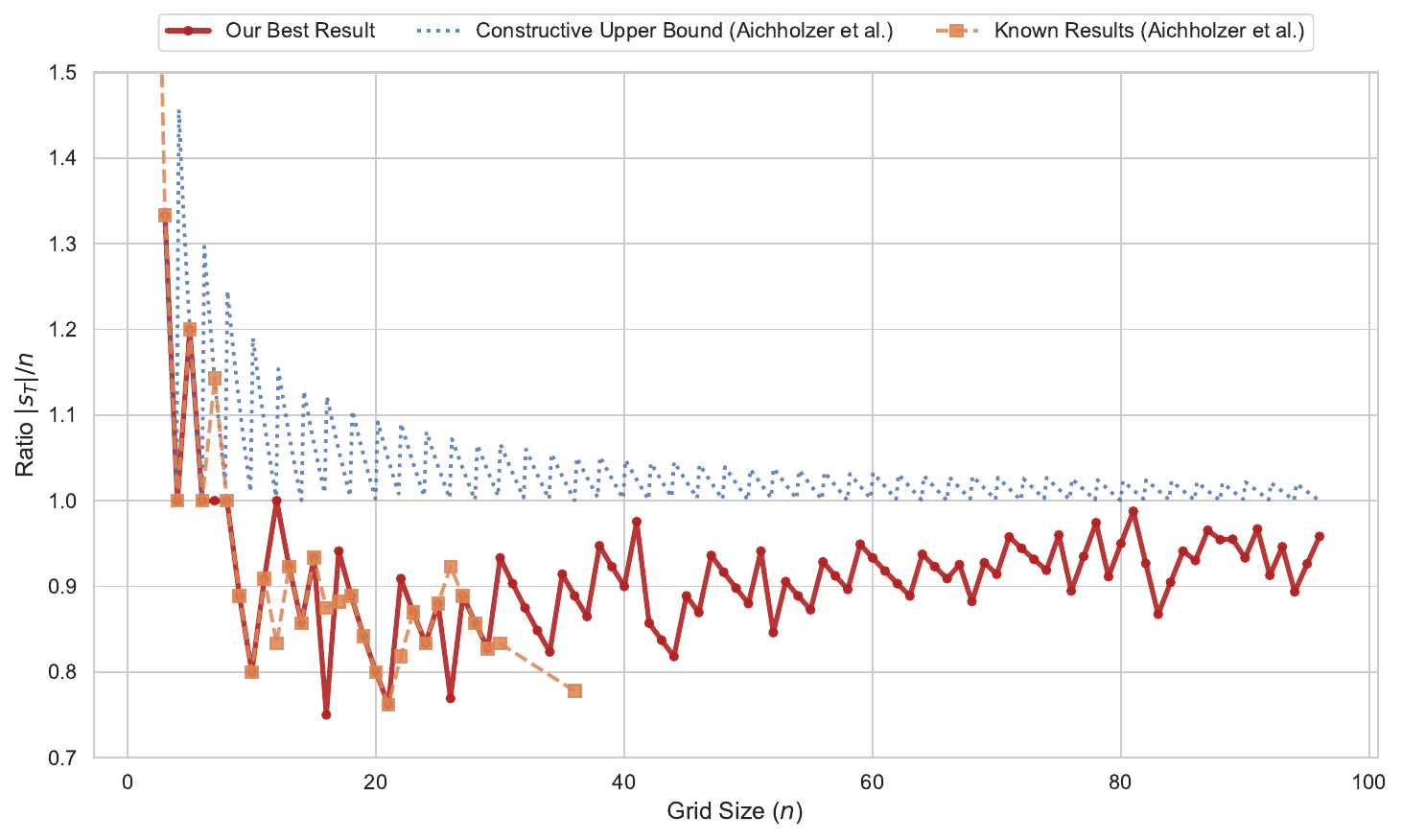}
        \caption{Min-Dom}
        \label{fig:res_min_dom}
    \end{subfigure}

    \vspace{0.5em}

    % --- Row 2 ---
    \begin{subfigure}[t]{0.32\textwidth}
        \centering
        \includegraphics[width=\linewidth]{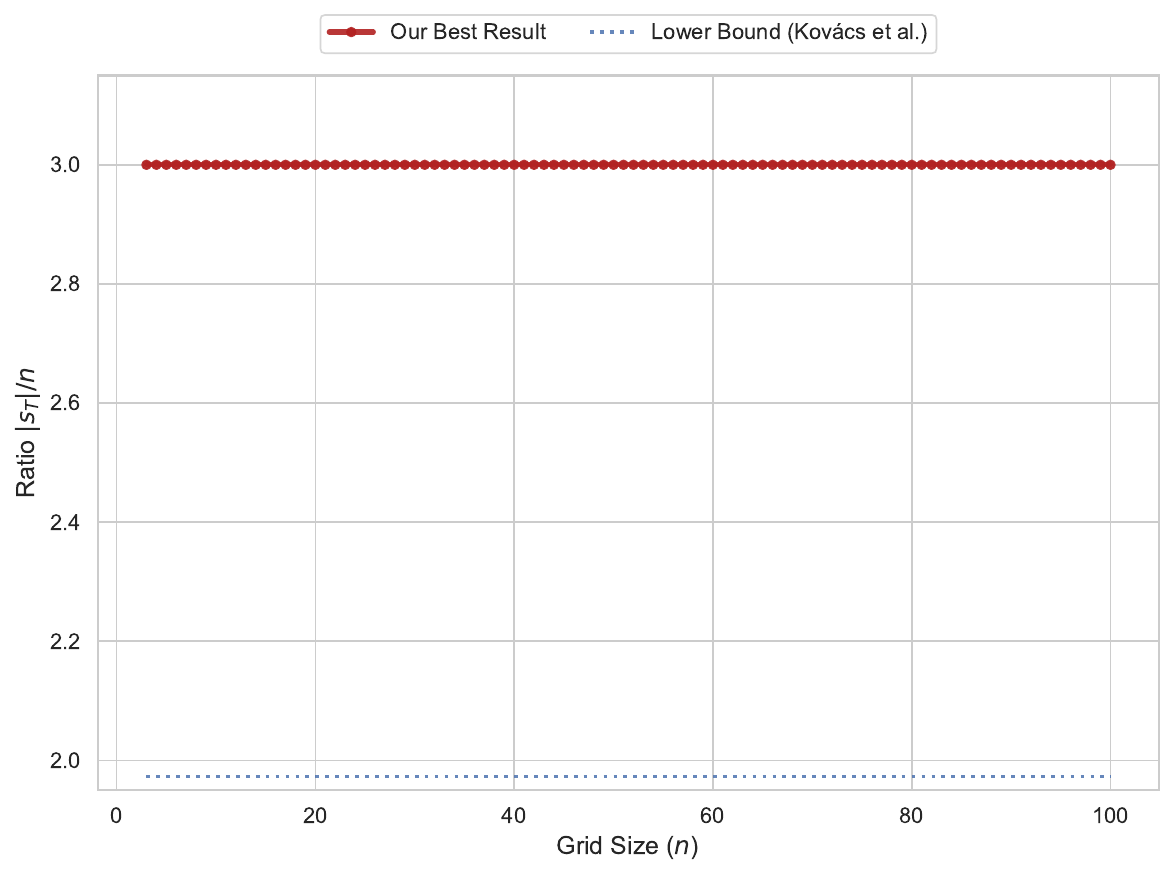}
        \caption{Max-N4IL}
        \label{fig:res_max_n4il}
    \end{subfigure}
    \hfill
    \begin{subfigure}[t]{0.32\textwidth}
        \centering
        \includegraphics[width=\linewidth]{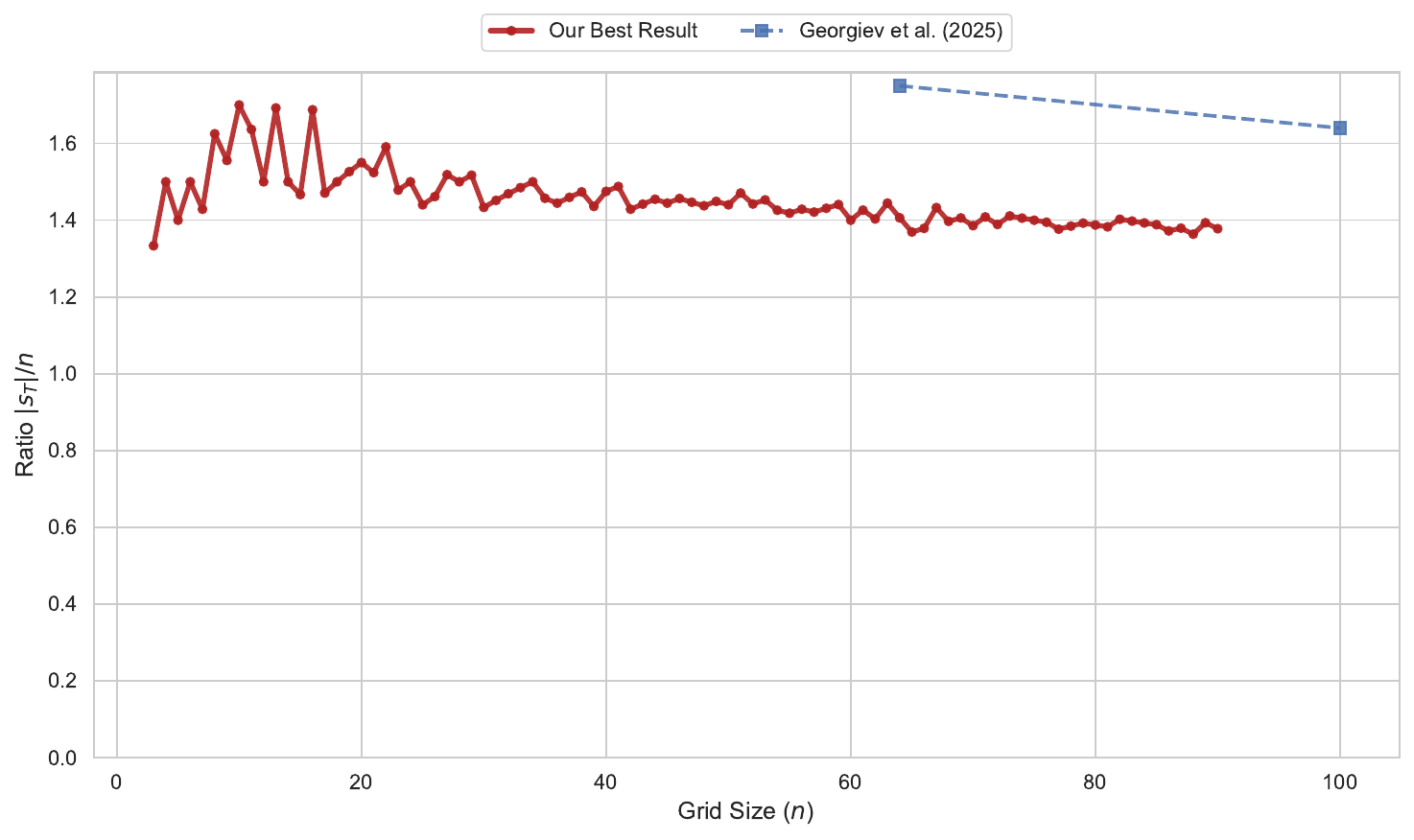}
        \caption{Max-No-Isosceles}
        \label{fig:res_no_isosceles}
    \end{subfigure}
    \hfill
    \begin{subfigure}[t]{0.32\textwidth}
        \centering
        \includegraphics[width=\linewidth]{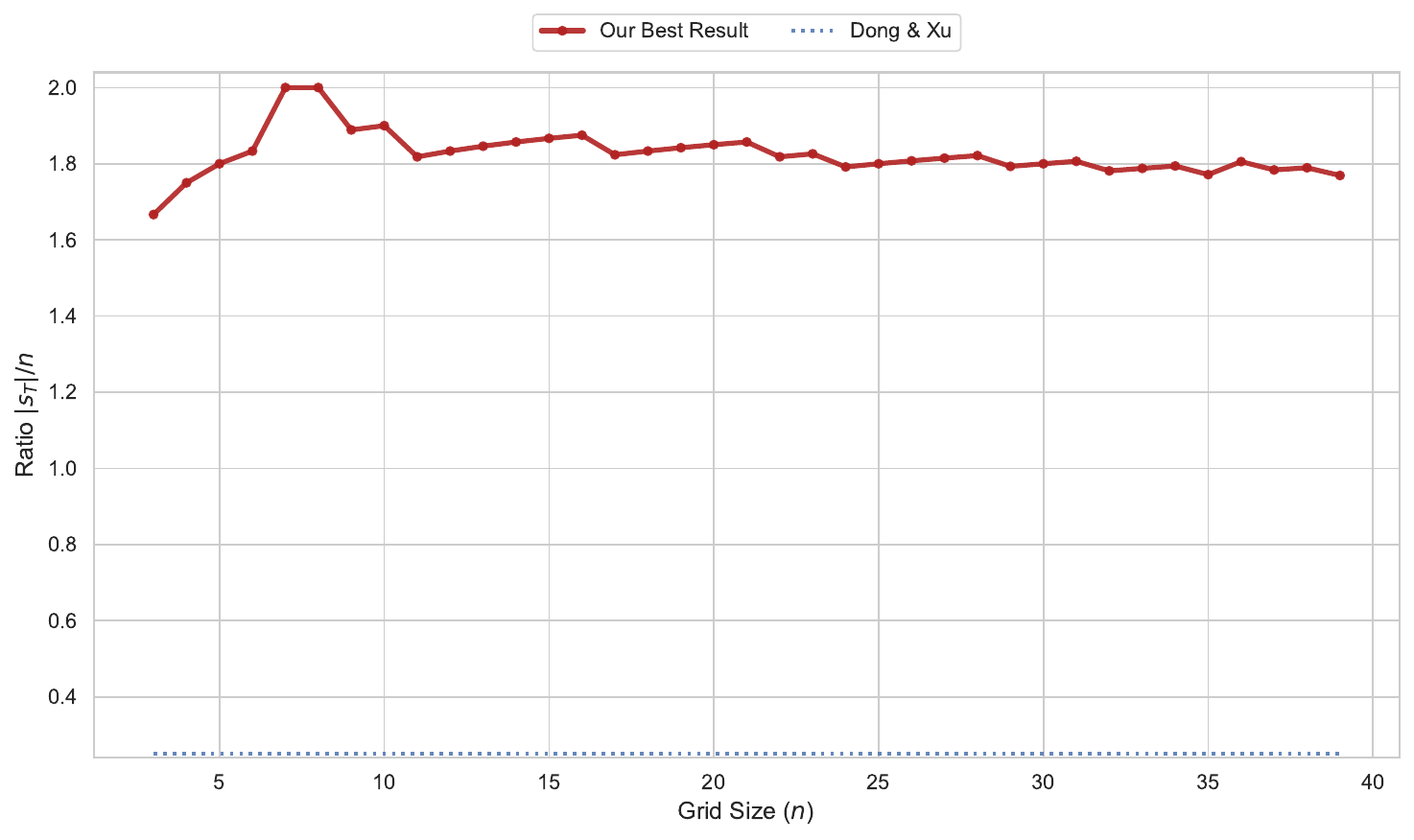}
        \caption{Max-No-4-on-Circle}
        \label{fig:res_no_4_on_circle}
    \end{subfigure}

    \caption{Scaling behaviors (optimality ratio $|s_T|/n$) across six constraints. Our results (red) are compared against previous best-known bounds. The framework successfully pushes empirical lower bounds upwards (\subref{fig:res_max_n3il}, \subref{fig:res_max_n4il}, \subref{fig:res_no_4_on_circle}) and tightens upper bounds downwards (\subref{fig:res_min_complete}, \subref{fig:res_min_dom}), demonstrating strong generalization across these kinds of geometric problems.}
    \label{fig:all_variants_scaling}
\end{figure}

The proposed framework establishes new best-known computational bounds for five out of the six problems we evaluated (\cref{tab:sota_comparison}). \Cref{fig:all_variants_scaling} illustrates the detailed scaling behaviors (optimality ratio $|s_T|/n$). While minor local variance exists due to MCTS stochasticity and opportunistic multi-trial sampling, the macroscopic trends robustly support our empirical bounds. The framework demonstrates exceptional search efficiency and generalizability across diverse constraints:

\textbf{Collinearity Constraints (\cref{fig:res_max_n3il,fig:res_max_n4il}):} Our MCTS framework finds sets of points of size roughly $1.8 n$ for each $n$ satisfying $\ourNothreeILnewbestknownbeginN \le n \le 119$.  This sits squarely between the algebraic lower bound of size approximately $1.5n$ and the upper bound of size $2n$.  This demonstrates the success of this method on grids that were previously inaccessible.

For the Max-N4IL problem, for each $n$ satisfying $3 \le n \le 100$ we were able to find optimal configurations with $3n$ points with no $4$ collinear points, outperforming the previous $1.973n$ lower bound. It would be interesting to see if we can produce these kinds of configurations for even larger grids.

\textbf{Coverage Objectives (\cref{fig:res_min_complete,fig:res_min_dom}):} For minimization tasks, our MCTS framework was most successful in finding  configurations with full $D_4$-symmetry.  We find complete sets of size approximately $0.95 n$ for each $n$ in the range $37 \le n \le 96$.  Since $\mathscr{D}_n \le \mathscr{I}_n$ for each $n$, this provides a new upper bound for both Min-Complete and Min-Dom in this range.  For some $n$ (e.g., $n=83$), we find a geometric dominating set of size smaller than the smallest complete set that we found.

\textbf{Non-linear \& Distance Constraints (\cref{fig:res_no_isosceles,fig:res_no_4_on_circle}):} We find configurations of approximately $1.8 n$ points with no $4$ points on a circle of finite radius for each $n \le 39$. For Max-No-Isosceles, the framework finds configurations of size approximately $1.4 n$. Highly specialized constructions using recent AI tools find larger sets for specific grid sizes (e.g., $n \in \{64,100\}$) \citep{georgiev2025mathematicalexplorationdiscoveryscale}, and could likely be applied to a whole range of values of $n$.  We still include the performance of our MCTS framework to demonstrate that it remains competitive with these more specialized approaches without requiring domain-specific heuristic redesign.  This establishes a boundary of our current search capabilities.

\subsection{Effectiveness of Strategies}
\label{subsec:effectiveness_strategies}

The effectiveness of symmetric batch transitions heavily depends on aligning the chosen subgroup $G$ with the geometric constraints of the problem. These geometric priors can be acquired through empirical exploration or existing mathematical literature.

\begin{figure}[ht]
    \centering
    % --- First Subfigure: Max-N3IL Batch Transitions ---
    \begin{subfigure}[t]{0.32\textwidth}
        \centering
        \includegraphics[width=\linewidth]{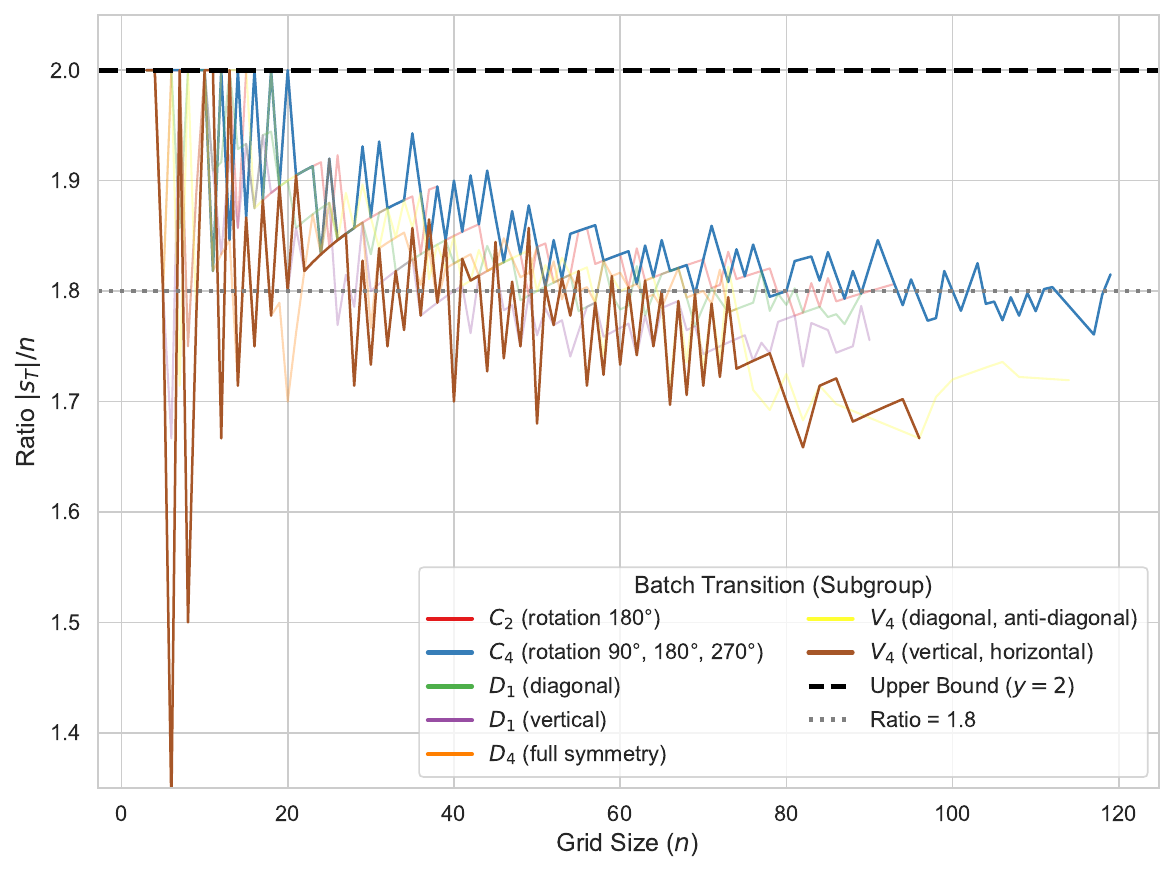}
        \caption{Max-N3IL Batch Transitions}
        \label{fig:batch_transition_max}
    \end{subfigure}
    \hfill
    % --- Second Subfigure: Min-Complete Batch Transitions ---
    \begin{subfigure}[t]{0.32\textwidth}
        \centering
        \includegraphics[width=\linewidth]{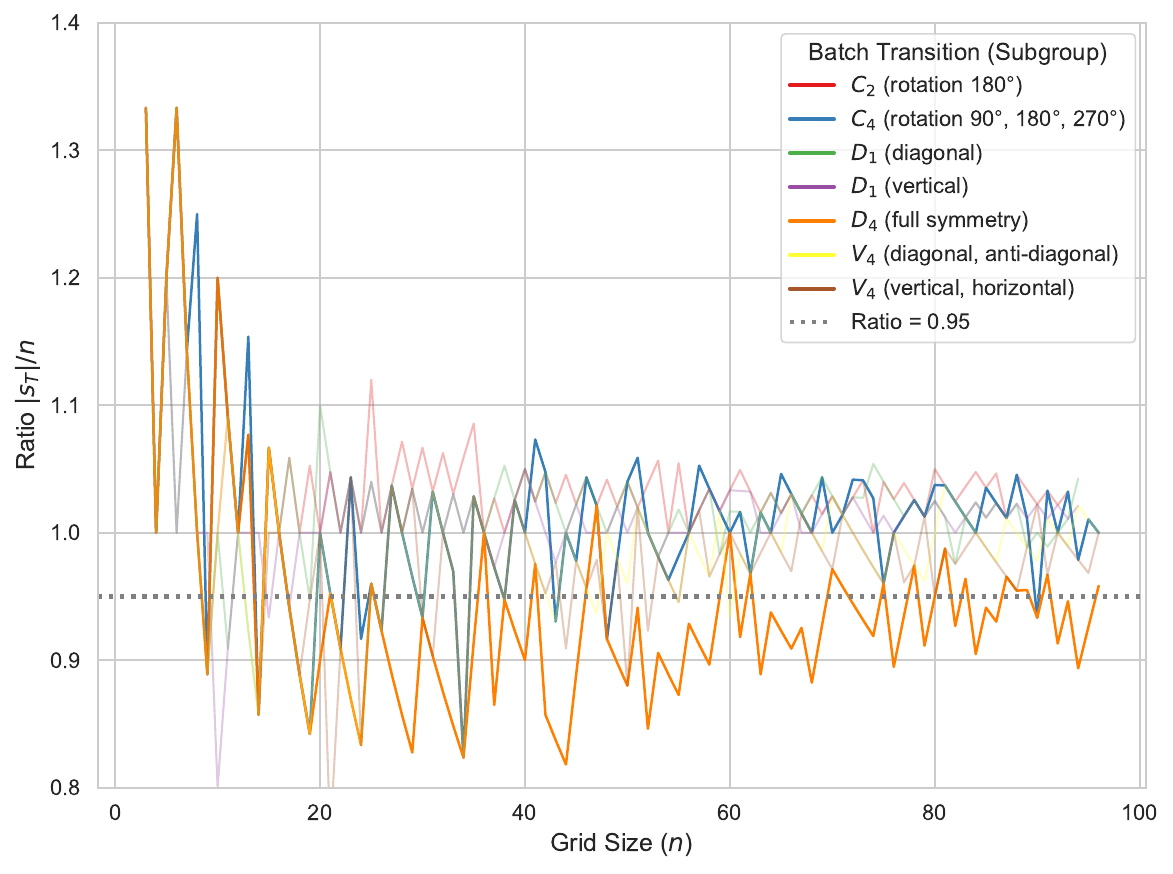}
        \caption{Min-Complete Batch Transitions}
        \label{fig:batch_transition_min}
    \end{subfigure}
    \hfill
    % --- Third Subfigure: Anytime Optimal Tracking ---
    \begin{subfigure}[t]{0.32\textwidth}
        \centering
        \includegraphics[width=\linewidth]{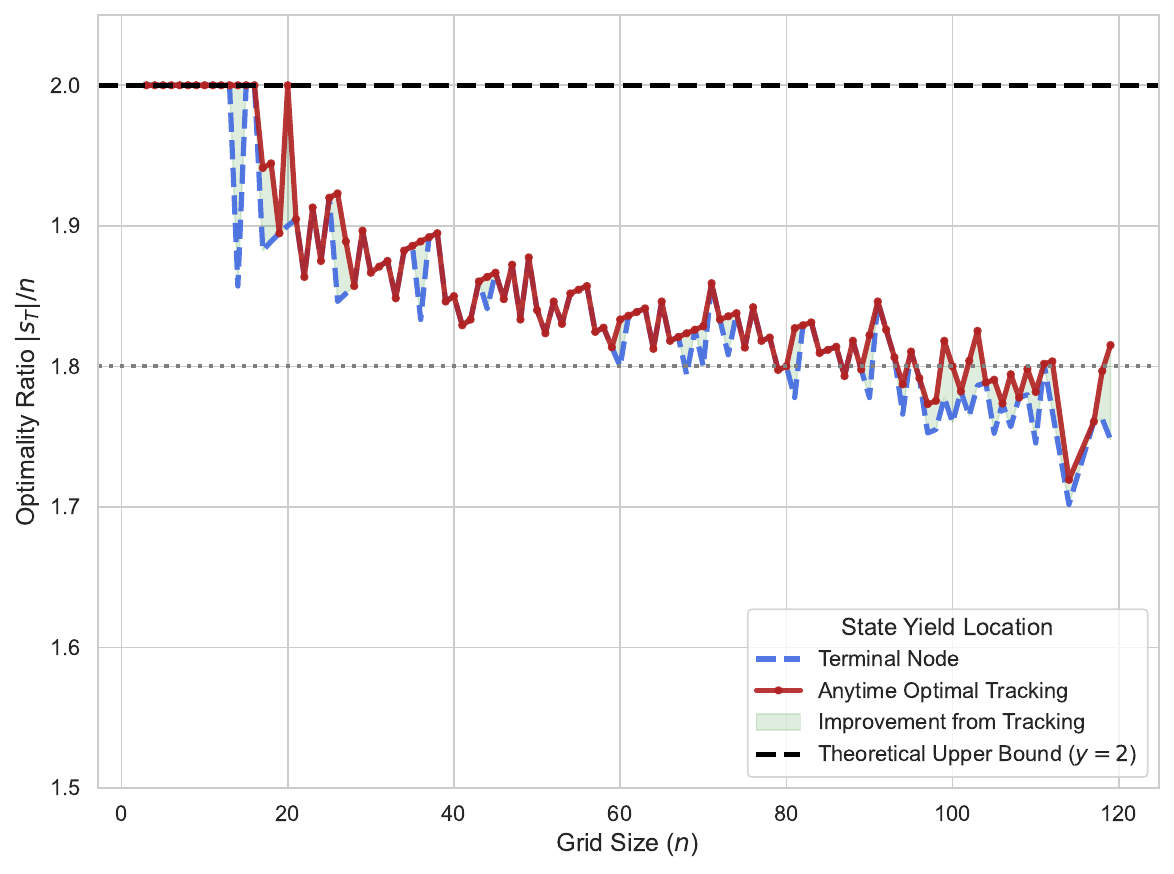}
        \caption{Anytime Optimal Tracking}
        \label{fig:anytime_tracking_gain}
    \end{subfigure}

    \caption{Effectiveness of algorithmic strategies. The optimal symmetric subgroup is heavily problem-dependent: $C_4$ excels for Max-N3IL (\subref{fig:batch_transition_max}) while $D_4$ excels for Min-Complete by rapidly covering the grid (\subref{fig:batch_transition_min}). Asynchronous tracking (\subref{fig:anytime_tracking_gain}) effectively captures outlier configurations discovered deep in rollouts.}
    \label{fig:effectiveness_batch_anytime}
\end{figure}

\textbf{Symmetric Batch Transitions.} For the Max-N3IL problem, empirical exploration (\cref{fig:batch_transition_max}) shows that the rotational subgroup $C_4$ yields the highest optimality ratios overall. Conversely, axis reflections ($V_4$) perform poorly, likely because they tend to rapidly remove whole rows and columns from $\mathcal{A}_{\text{feas}}$. However, this exact rapid-elimination mechanism becomes a powerful advantage for minimization tasks. For the Min-Complete problem (\cref{fig:batch_transition_min}), the full symmetry group $D_4$ leverages these reflections to rapidly cover the grid, successfully driving the search toward sparser dominating sets. Furthermore, the literature confirms that known optimal N3IL configurations often exhibit $C_4$ or diagonal symmetries \citep{flammenkamp1992,flammenkamp1998,flammenkamp_no3in_website}, motivating our specific subgroup selections for large grids ($n \ge 96$). If symmetric placements prove detrimental to a specific problem, the framework adapts to a single-action search.

\textbf{Anytime Optimal Tracking.} \Cref{fig:anytime_tracking_gain} demonstrates the value of asynchronously recording the global best state during random simulations. This tracking permanently harvests valid ``outlier'' solutions that might otherwise be discarded when trees are reset or pruned, ensuring the final output is strictly no worse than terminal node evaluations.

We analyze the incremental contribution of each algorithmic component in our ablation study (Section \ref{subsec:ablation_study}).

\subsection{Ablation Study}

\label{subsec:ablation_study}

\begin{figure}[hbt]
    \centering
    \includegraphics[width=0.7\linewidth]{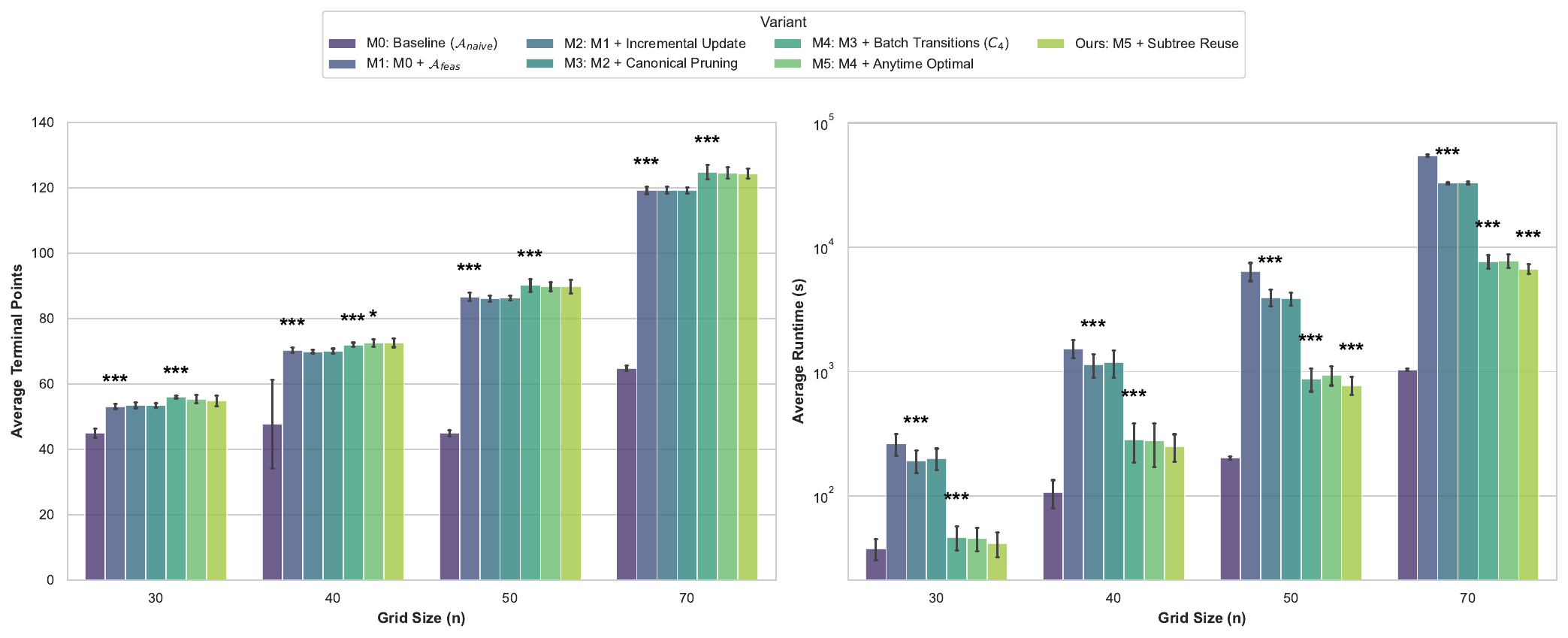}
    \caption{Ablation Study: Comparison of average terminal points (left) and average runtime (right) across different algorithm variants and grid sizes.
    Bar heights represent the mean value over $N=20$ independent random seeds, and error bars indicate the standard deviation ($\pm 1$ std).
    Statistical significance is evaluated incrementally against the immediately preceding variant using a one-sided Welch's t-test (* $p < 0.05$, ** $p < 0.01$, *** $p < 0.001$).
    Specifically, the alternative hypothesis for terminal points is that the current variant yields a strictly greater mean ($\mu_{\text{curr}} > \mu_{\text{prev}}$), whereas for runtime, the alternative hypothesis is a strictly lower mean ($\mu_{\text{curr}} < \mu_{\text{prev}}$).
    Note that runtime is plotted on a logarithmic scale. See \cref{tab:incremental_ablation} for detailed numerical results.}
    \label{fig:ablation_study_combined}
\end{figure}

The different components of our framework exhibit hierarchical dependencies; for instance, \textit{Canonical Pruning} relies on the structure of the \textit{Feasible Action Space} ($\mathcal{A}_\text{feas}$), and \textit{Symmetric Batch Transitions} build directly upon this pruning mechanism. Therefore, we start with the baseline MCTS and cumulatively add components to evaluate the marginal performance gain of each layer.

As shown in \cref{fig:ablation_study_combined} and \cref{tab:incremental_ablation}, our final variant achieves superior performance in producing successful configurations and also demonstrates highly optimized computational efficiency. While minor runtime fluctuations exist---likely attributable to the stochastic nature of the shared cluster environment---the overall improvements remain robust across all tested grid sizes.

The first major performance leap occurs with the strict enforcement of the feasible action space (\textbf{M1}). Compared to the baseline (\textbf{M0}), which operates on the naive action space $\mathcal{A}_\text{naive}$ and frequently terminates early due to invalid random selections, M1 exhibits a significant improvement in average terminal number of points ($p<0.001$ across all grid sizes) by ensuring the deep exploration of valid configurations. However, computing $\mathcal{A}_\text{feas}$ from scratch at each node severely degrades runtime. This bottleneck is resolved by the incremental update mechanism (\textbf{M2}), which dramatically improves runtime compared to M1 ($p<0.001$) while incurring no statistically significant performance loss ($p \ge 0.05$). This efficiency gain scales exceptionally well for larger grids, such as $n \in \{50, 70\}$, empirically validating the theoretical $O(n^2)$ update complexity.

The subsequent integration of canonical pruning (\textbf{M3}) yields no statistically significant improvement in either performance or efficiency on its own. Without forced symmetric transitions, symmetric states become increasingly sparse at deeper search levels. Consequently, the computational overhead of computing the stabilizer at each step outweighs the minimal branching reduction achieved. This limitation is directly overcome by the introduction of $C_4$ symmetric batch transitions (\textbf{M4}), which creates a breakthrough in both metrics. By actively maintaining structural symmetry deeper within the search tree, the batch mechanism unlocks the potential of canonical pruning. As a result, M4 establishes new empirical peaks for terminal points ($p<0.001$) and drastically reduces runtime compared to the M1--M3 variants.

Finally, we evaluate the utility-focused enhancements. Although the anytime optimal tracking (\textbf{M5}) shows no statistically significant shift in average performance within this specific randomized ablation study, it serves as a critical, zero-cost fail-safe. It ensures that the globally best configuration discovered during any rollout is permanently recorded, independent of the final MCTS policy convergence. The framework is finalized with the subtree reuse strategy (\textbf{Ours}), which significantly improves overall efficiency on larger grids ($p<0.001$ for $n \in \{50, 70\}$) with no statistically significant loss in terminal points. This confirms that persisting tree statistics across deterministic transitions effectively reduces redundant explorations, demonstrating the scalability of the framework for a wide array of extremal problems in combinatorial geometry.

\section{Conclusions and Limitations}
\label{sec:conclusion}

In this work, we presented a Geometry-Aware Monte Carlo Tree Search framework tailored for extremal problems with hard geometric constraints. By integrating \textit{Incremental Feasible Action Space} maintenance and leveraging symmetries through \textit{Canonical Pruning} and \textit{Symmetric Batch Transitions}, we help to address the ``validity cliff'' and combinatorial explosion that hinder generic solvers.

Our empirical results demonstrate the effectiveness of this approach. The MCTS framework we propose achieves a significant improvement in performance over baseline implementations.  Our ablation study shows how the various aspects of this approach lead to improved efficiency.  We establish new best-known bounds for five of the six problems within the grid sizes we evaluated.  This demonstrates the capability of our method to generalize and scale, suggesting that it can serve as a versatile tool for constructions in discrete geometry.

Despite the combinatorial explosion inherent to these problems, our framework demonstrates remarkable scalability. All reported main results, including the exploration of grids up to $n = 119$ for Max-N3IL, were achieved using only a \textbf{single CPU core with a strict 6GB memory limit} within a two-week cutoff. This confirms that our geometry-aware pruning effectively mitigates the curse of dimensionality for a significant range of $n$.

Our method provides empirical constructions rather than closed-form bounds. The configurations that we produce are not necessarily optimal, but still provide valuable information for grids of larger sizes that could be used to guide future theoretical work. Future work will leverage parallelization and GPU acceleration \citep{KLESK2025102139} to explore even larger spaces. Integrating neural networks for value/policy guidance for this geometry-aware search is a promising direction for future research.

\subsubsection*{Acknowledgments}

We acknowledge support from NSF Grant DMS 2154223.  L. Zhang was supported by a UCI Undergraduate Research Opportunities Program (UROP) Summer Fellowship. 

We thank Jeremy Jiang, Desmond West-Hedlund, Miguel Angel Estrella-Ibarra, and Erin Hanrahan, for stimulating discussions. We also thank Hengrui Cai, David Eppstein, and Wenbo Zhang for valuable feedback and suggestions.  Computational resources (mentioned in Section \ref{subsec:exp_setup}) were provided by the High Performance Community Computing Cluster (HPC3) at UCI\footnote{\url{https://rcic.uci.edu/hpc3/hpc3.html}}. We thank Ge Gao and the HPC3 Support Team for technical support. 

\bibliography{main}
\bibliographystyle{plainnat}

\appendix

\section{Related Works}
\label{sec:related_works}

We categorize existing approaches to the extremal problems into two distinct streams: theoretical mathematical constructions and constrained tree search.

\subsection{Mathematical Bounds and Constructions}

\subsubsection{No-3-in-Line}
Early work focused on number-theoretic constructions to establish foundational lower bounds. \citet{hall1975} utilized points on modular hyperbolas ($xy \equiv k \pmod p$) to achieve the current best asymptotic lower bound of roughly $1.5n$. However, a significant gap remains between this constructive lower bound and the trivial upper bound of $2n$. This substantial margin suggests that for finite grid sizes (e.g., $n \in [50, 100]$), $M(n)$---the maximum number of points in an $n \times n$ grid such that no $3$ points are collinear---may be significantly higher than what algebraic constructions can guarantee. Motivated by this gap, our work employs computational search to explore the solution space within this interval, aiming to empirically push the lower bound upward and reveal larger configurations than the ones that we can produce with algebraic constructions.

\subsubsection{Geometric Dominating Sets and Independent Geometric Dominating Sets}

Following \citet{AICHHOLZER2023101913}, a set $S\subseteq \mathcal{G}_n$ is a
\emph{geometric dominating set} if every grid point $p$ is either in $S$ or lies on a line
determined by two points of $S$; equivalently, every $p$ is collinear with two (distinct) points
of $S$.  The minimum size of such a set is the \emph{geometric domination number} $\mathscr{D}_n$. If additionally no three points of $S$ are collinear, then $S$ is an
\emph{independent geometric dominating set}. Let $\mathscr{I}_n$ denote the minimum size of a independent geometric dominating set in an $n \times n$ grid. The authors also establish the first general asymptotic bounds for $\mathscr{D}_n$:
they prove a lower bound $\mathscr{D}_n=\Omega(n^{2/3})$ and give an explicit construction of a geometric dominating set of size $2 \lceil n/2\rceil$. These results leave a substantial gap between the best known lower and upper bounds
in the grid, motivating both improved constructions and stronger counting/packing
arguments.

We understand much less about the minimum size of an independent geometric dominating set.  It is clear that $\mathscr{I}_n \le 2n$, but we do not even know that there exist infinitely many $n$ for which $\mathscr{I}_n < 2n$.

On the computational side, \citet{AICHHOLZER2023101913} compute the value of $\mathscr{I}_n$ for $n \le 12$ and give upper bounds for $n$ satisfying $13 \le n \le 30$ and $n=36$, providing evidence and test cases for conjectures
about the true growth of $\mathscr{I}_n$.%

\subsubsection{No-4-in-Line}
Let \( f_{4\text{-coll}}(n) \) denote the maximum cardinality of a subset of \( \mathcal{G}_n \) containing no four collinear points. The best asymptotic bounds currently known, established by \citet{kovacs2025randomised}, are:
\[
1.973\,n \le f_{4\text{-coll}}(n) \le 3n
\]
for all sufficiently large \( n \). Consequently, while the maximum size is known to scale linearly with \( n \),  we do not even have a conjecture for how $f_{4\text{-coll}}(n)$ grows with $n$.

\subsubsection{No-4-on-a-Circle}

Let \( f_{\mathrm{circ}}(n) \) denote the maximum cardinality of a subset of \( [n]^2 \) containing no four concyclic points. The tightest asymptotic bounds currently known, established by \citet{dong2025largegridsubsetscospherical}, are:
\[
n-o(n) \le f_{\mathrm{circ}}(n) \le \frac{5}{2}n - \frac{3}{2}.
\]
Consequently, while the maximum size is guaranteed to grow linearly with \( n \), we do not even have a conjecture for how $f_{\mathrm{circ}}(n)$ grows with $n$.

\subsubsection{No-Isosceles-Triangle}

Let \(f_{\mathrm{iso}}(n)\) denote the maximum size of a subset of \([n]^2\) containing no three points that form an isosceles triangle, including degenerate collinear configurations.  The best known lower bound is:
\[
f_{\mathrm{iso}}(n)\ge c\,\frac{n}{\sqrt{\log n}}
\]
for some absolute constant \(c>0\) \citep{charton2024patternboostconstructionsmathematicslittle}.

The best known upper bounds for this problem come from bounding the number
of points on each horizontal line of the grid: three points at positions
$(a,y)$, $(a+d,y)$, $(a+2d,y)$ on the same horizontal line form a
(degenerate) isosceles triangle, so the restriction of any isosceles-free
set to a single horizontal line must avoid three-term arithmetic
progressions.  Applying the best known bounds for subsets of
$\{1,2,\ldots,n\}$ containing no three-term arithmetic progression to each
of the $n$ horizontal lines, \citet{bloom2023improvementkelleymekaboundsthreeterm}
established
\[
f_{\mathrm{iso}}(n)\le \mathrm{exp}(-O((\log n)^{1/9}))\,n^2.
\]
This was further improved by \citet{raghavan2026}, giving

\[
f_{\mathrm{iso}}(n) \le \mathrm{exp}({-O((\log n)^{1/6}(\log\log n)^{-1}})) \, n^2.
\]

Like in the problems about sizes of dominating sets, the upper and lower bounds here have different orders of magnitude.

\subsection{Monte Carlo Tree Search}

Monte Carlo Tree Search (MCTS) was introduced by \citet{coulom2006efficient} as a framework combining selective tree search with random rollout evaluations. The UCT algorithm of \citet{kocsis2006bandit} provided a principled exploration-exploitation strategy by treating node selection as a multi-armed bandit problem, and remains the backbone of most modern MCTS implementations \citep{browne2012survey}. MCTS gained widespread attention through AlphaGo and AlphaZero \citep{silver2016mastering, silver2018general}, which demonstrated superhuman performance on Go, Chess, and Shogi by coupling MCTS with deep neural networks. These works also established engineering practices---most notably subtree reuse across decision steps---that we adopt in our framework. However, neural networks are not included in our current implementation, and we leave the integration of learned value and policy functions to future work.

Beyond game playing, MCTS has been applied to combinatorial optimization problems such as scheduling and vehicle routing \citep{jooken2022mcts}. However, these settings typically feature smooth objective functions and separable constraints, where random rollouts naturally produce valid solutions. The geometric constraints in this work (e.g., collinearity, concyclicity) are \textit{global}: a single point placement invalidates an $O(n)$ set of grid cells, and uniform random rollouts over the unrestricted action space collapse with high probability. This incompatibility motivates our Incremental $\mathcal{A}_\text{feas}$ mechanism, which maintains geometric validity throughout the search at $O(n^2)$ cost per step.

Two prior lines of work address symmetry within MCTS. \citet{schiffel2010symmetry} exploited symmetry in general game playing (GGP) by detecting automorphisms at the \textit{game rule level} and merging statistics of globally equivalent states via a transposition table. \citet{cazenave2025mcps} proposed Monte Carlo Permutation Search, which fuses visit statistics of permutation-equivalent actions to reduce effective branching. Our approach differs from both in two respects. First, rather than detecting global game symmetry or fusing statistics, we dynamically compute the \textit{state-level stabilizer subgroup} $\text{Stab}(s) \leq D_4$ at each node and restrict expansion to canonical orbit representatives---a finer-grained, state-dependent pruning. Second, our Symmetric Batch Transitions redefine the MDP transition itself by attempting simultaneous orbit-wide point placements, a mechanism with no direct counterpart in prior MCTS literature.

The challenge of tuning the exploration--exploitation balance over a long horizon has been addressed by SA-UCT \citep{ruijl2013combiningsimulatedannealingmonte}, which anneals the UCB exploration constant over iterations. We adopt a similar decay schedule but introduce a refinement for the subtree reuse setting: our progress ratio $r$ incorporates inherited visit counts $N_{\text{existing}}$, ensuring that the exploration schedule remains continuous and monotonically decreasing across decision steps rather than resetting at the start of each move.

\section{Computational Complexity Analysis}
\label{appendix:complexity}

We analyze the time complexity of our Geometry-Aware MCTS framework, using the Max-N3IL problem on the $n \times n$ grid as a representative case. We first derive the per-iteration cost, then aggregate over the full search to establish the total complexity, and finally contrast this polynomial bound with the intractability of exact methods.

\paragraph{Per-Iteration Cost Breakdown.}
Each MCTS iteration consists of four phases. We analyze their worst-case costs in terms of $n$:

\begin{description}
    \item[Selection.] The agent traverses the tree from root to leaf. Since each level corresponds to a point placement and the trajectory length is bounded by $2n$, the tree depth is $O(n)$. At each internal node, the UCB1 score is evaluated over all expanded children, costing $O(|\mathcal{A}_\text{feas}|)$ where $|\mathcal{A}_\text{feas}|$ is equal to the local branching factor. In the worst case $|\mathcal{A}_\text{feas}| = O(n^2)$, but in practice the feasible action space shrinks rapidly with depth (cf.~Section~\ref{subsec:incremental_svas}). The total selection cost is $O(n) \cdot O(|\mathcal{A}_\text{feas}|)$.

    \item[Expansion.] A single child node is created. The dominant cost is the incremental update of the feasible action space via ray casting (Section~\ref{subsec:incremental_svas}): the new point forms lines with each of the $|s| \le 2n$ existing points, and each line is traced across $O(n)$ grid cells. This yields an expansion cost of $O(n^2)$ per node.

    \item[Simulation (Rollout).] Starting from the expanded node, the agent performs a random rollout until the terminal condition $\mathcal{A}_{\text{feas}} = \emptyset$ is reached. Each rollout step selects a random feasible action and incrementally updates $\mathcal{A}_{\text{feas}}$ at a cost of $O(n^2)$. Since at most $2n$ points can be placed in total, the rollout traverses at most $O(n)$ steps. The simulation cost is therefore $O(n) \cdot O(n^2) = O(n^3)$.

    \item[Backpropagation.] The reward is propagated from the leaf back to the root. Each update is $O(1)$, and the path length is $O(n)$, giving a total cost of $O(n)$.
\end{description}

The simulation phase dominates, yielding a \textbf{per-iteration cost of $O(n^3)$}.

\paragraph{Total Complexity.}
The full algorithm proceeds through two nested loops. The \textit{outer loop} executes the sequential MDP: at each decision step, MCTS selects and commits to an action, advancing the trajectory by one point. Since the trajectory length is at most $2n$, the outer loop runs for $O(n)$ steps. The \textit{inner loop} performs MCTS search at each decision step, executing $N_{\text{iter}} = O(n^2)$ iterations (we set $N_{\text{iter}} = O(n^2)$ in all experiments). Combining with the per-iteration cost:
\begin{equation}
    T_{\text{total}} = \underbrace{O(n)}_{\text{trajectory}} \times \underbrace{O(n^2)}_{\text{iterations per step}} \times \underbrace{O(n^3)}_{\text{cost per iteration}} = O(n^6).
\end{equation}

The total time complexity of our framework is therefore \textbf{polynomial} in the grid size $n$.

\paragraph{Contrast with Exact Methods.}
The polynomial complexity of our approach stands in sharp contrast to the cost of exact enumeration. An exhaustive search must, in principle, examine all candidate subsets of the $n^2$ grid points up to the trivial upper bound of $2n$. The size of this search space is
\begin{equation}
    \sum_{k=1}^{2n} \binom{n^2}{k},
\end{equation}
which grows super-exponentially in $n$. Therefore, exact methods are severely limited for large grid sizes.

\paragraph{The Tractability--Optimality Trade-off.}
Our framework does not guarantee global optimality; it is a heuristic search that trades optimality guarantees for polynomial-time tractability. However, this trade-off is precisely what enables the exploration of grid sizes $(n> 68)$ that up to this point have been too large for exact computation. As demonstrated in Section~\ref{sec:method}, the components of our geometry-aware framework---incremental $\mathcal{A}_\text{feas}$ maintenance, canonical pruning, and symmetric batch transitions---ensure that this polynomial budget is spent efficiently, concentrating search effort on the most promising regions of the combinatorial landscape. The empirical results (Section~\ref{sec:results}) confirm that this approach yields configurations that significantly exceed the best-known constructive lower bound of $\approx 1.5n$ and closely approach the conjectured density of $\approx 1.814n$ \citep{guy1968}, suggesting that $O(n^6)$ may be capable of identifying structures that approach the current conjectured bounds.

\paragraph{Generality Across Problem Variants.}
The $O(n^3)$ per-iteration bottleneck arises from the incremental maintenance of $\mathcal{A}_\text{feas}$ during simulation, which is specific to the geometric invariant $\Phi$. For other problem variants (e.g., No-Four-in-Line, No-Four-on-a-Circle), the per-step update cost may differ depending on the constraint structure, but the overall algorithmic skeleton---and its polynomial-time guarantee---remains unchanged.

\section{Detailed Experimental Configurations}
\label{appendix:detailed_exp_configurations}

We list the detailed hyperparameters used in our experiments to ensure reproducibility. Table~\ref{tab:hyperparams} summarizes the configurations for the MCTS algorithm. The UCT algorithm specified in Section \ref{subsubsec:exploration_decay} was utilized for all experiments.

\subsection{Reward Functions}
\label{subsec:reward_function}

For every extremal problem with the goal of maximizing $|s_T|$ and minimizing $|s_T|$, we use these two reward functions respectively:
\begin{equation}
    R(s_T, n) = \exp{\left({2 \frac{|{s_T}|-n}{n}}\right)} \quad \text{and} \quad R(s_T, n) = \exp{\left({-2 \frac{|{s_T}|-n}{n}}\right)}.
\end{equation}
The design of the reward function is an empirical choice aimed at amplifying signal when the optimality gap is small. Optimizing the reward shaping is orthogonal to the main contribution of this work.

\subsection{Hyperparameter Settings for Main Experiments}

We set these hyperparameters and configurations based on complexity, best-known geometric prior, and heuristics for each problem (see \Cref{tab:hyperparams}).

\begin{table}[ht]
    \centering
    \small %  resizebox
    \caption{Detailed hyperparameter settings and algorithmic configurations across problems.}
    \label{tab:hyperparams}
    \setlength{\tabcolsep}{4pt}
    \begin{tabular}{l c l c c}
        \toprule
        \textbf{Hyperparameters} & \textbf{Max Iterations} & \textbf{Batch Transition} & \textbf{Max Depth for} & \textbf{Exploration} \\
        \textbf{Symbol} & $N_{\text{iter}}$ & $G$ & \textbf{Canonical Pruning} & \textbf{Constant} $C$ \\
        \midrule
        \textbf{Max-N3IL} ($3 \leq n \leq 95$) & $100n^2$ & 1 trial of each symmetry & \multirow{4}{*}{$2n$} & \multirow{7}{*}{$1.41$} \\
        \cmidrule{1-3}
        \textbf{Max-N3IL} ($96 \leq n \leq 119$) & $25n^2$ & 1 trial each in $C_4$ and $V_4$\textsuperscript{(a)} & & \\
        \cmidrule{1-3}
        \textbf{Min-Complete} & $100n^2$ & 1 trial of each symmetry & & \\
        \cmidrule{1-3}
        \textbf{Min-Dom} & $100n^2$ & 1 trial of each symmetry & & \\
        \cmidrule{1-4}
        \textbf{Max-N4IL} & $10n^2$ \& $25n^2$\textsuperscript{(b)} & $C_4$ & $3n$&  \\
        \cmidrule{1-4}
        \textbf{Max-No-Isosceles} & $100n^2$ & None & \multirow{2}{*}{$2$} & \\
        \cmidrule{1-3}
        \textbf{Max-No-4-on-Circle} & $25n^2$ & None & & \\
        \bottomrule
        \multicolumn{5}{l}{\footnotesize \textsuperscript{(a)} $V_4$ here specifically denotes diagonal and anti-diagonal reflections.} \\
        \multicolumn{5}{p{\textwidth}}{\footnotesize \textsuperscript{(b)} We ran 1 trial for each value of max iteration. \Cref{fig:res_max_n4il} shows the max number of points found for each $n$.}
    \end{tabular}
\end{table}

\subsection{Ablation Study Configurations}
\label{subsec:hyperparams_ablation}

The ablation study was conducted on the Max-N3IL problem. Table~\ref{tab:ablation_config} details the algorithmic configurations for each variant evaluated in the incremental ablation study.

\begin{table}[hbt]
    \centering
    \caption{Algorithmic configurations for each variant in the incremental ablation study. ``Max Depth'' indicates the maximum search depth up to which canonical pruning is applied. $2n$ represents full-depth pruning, as the maximum possible number of points is bounded by $2n$ for Max-N3IL. ``Opt. Tracking'' represents anytime optimal tracking.}
    \label{tab:ablation_config}
    \resizebox{\textwidth}{!}{%
    \begin{tabular}{llccccc}
        \toprule
        \textbf{Variant} & \textbf{Description} & \textbf{Action Space} & \textbf{Max Depth} & \textbf{Batch Transitions} & \textbf{Opt. Tracking} & \textbf{Tree Reuse} \\
        \midrule
        \textbf{M0} & Baseline & $\mathcal{A}_\text{naive}$ & None & None & Disabled & Disabled \\
        \textbf{M1} & M0 + $\mathcal{A}_\text{feas}$ & $\mathcal{A}_\text{feas}$ (Scratch) & None & None & Disabled & Disabled \\
        \textbf{M2} & M1 + Incremental & $\mathcal{A}_\text{feas}$ (Incremental) & None & None & Disabled & Disabled \\
        \textbf{M3} & M2 + Canonical Pruning & $\mathcal{A}_\text{feas}$ (Incremental) & $2n$ & None & Disabled & Disabled \\
        \textbf{M4} & M3 + Batch Transitions & $\mathcal{A}_\text{feas}$ (Incremental) & $2n$ & Rotational ($C_4$) & Disabled & Disabled \\
        \textbf{M5} & M4 + Anytime Optimal & $\mathcal{A}_\text{feas}$ (Incremental) & $2n$ & Rotational ($C_4$) & Enabled & Disabled \\
        \textbf{Ours} & M5 + Subtree Reuse & $\mathcal{A}_\text{feas}$ (Incremental) & $2n$ & Rotational ($C_4$) & Enabled & Enabled \\
        \bottomrule
    \end{tabular}%
    }
\end{table}

\paragraph{Other Settings for Ablation Study:}
The computational budget and evaluation metrics were strictly controlled across all variants. The experiments were conducted on grid sizes $n \in \{30, 40, 50, 70\}$. For each algorithmic variant and grid size, we evaluated the performance over 20 independent trials, initialized with random seeds $0$ through $19$. The search budget ($N_{iter}$) was uniformly set to $10n^2$ iterations per decision step.

\newpage

\section{Technical Details}
\label{appendix:tech_details}

\textbf{Memory Efficiency and State Space Complexity:}
    As stated in the main text, a strict 6GB memory limit was intentionally imposed across all trials to rigorously test the memory efficiency of our framework. The underlying state space complexity $|\mathcal{S}| = 2^{n^2}$ grows explosively; for instance, a mere $20 \times 20$ grid already contains $|\mathcal{S}| = 2^{400} \approx 2.58 \times 10^{120}$ states. By achieving superior results on significantly larger grids (e.g., up to $n=119$) within this modest memory budget, we empirically highlight the effectiveness of our geometry-aware pruning and incremental update mechanisms in mitigating the curse of dimensionality.

\textbf{Hardware Heterogeneity and Control:}
    The HPC cluster is a heterogeneous environment comprising various CPU architectures (e.g., Intel Skylake, Cascade Lake, and Ice Lake). To ensure the scientific rigor of our \textbf{ablation studies}, we enforced strict hardware consistency by using Slurm constraints to the best of our ability. Specifically, all ablation trials were restricted to nodes supporting the following features:
    \texttt{intel, avx512, mlx5\_ib, nvme, fastscratch}.
    This ensures that performance variations are attributable to algorithmic changes rather than underlying hardware differences.

\textbf{Resource Optimization for Main Experiments:}
    For non-ablation experiments (e.g., searching for maximum $n$ sets), where the primary goal is sheer computational throughput rather than controlled timing comparisons, we did not impose these hardware constraints. This allowed the scheduler to utilize any available CPU-only nodes, thereby maximizing resource utilization and reducing queue wait times under the 7-day or 14-day time limits.

\clearpage

\section{Detailed Previous Best-Known Results of Max-N3IL and Min-Complete Sets}

\label{appendix:detailed_previous_best_known}

\paragraph{Previous Best-Known Results of Largest No-Three-in-Line Sets:} \Cref{tab:best_known_n3il_detailed} shows the best detailed results known results according to \citet{flammenkamp_no3in_website, prellberg2026constraintsatisfactionprogrammingnothreeinline}.

\begin{table}[h]
\centering
\caption{Previous best known values for the sizes of the N3IL sets ($n \leq 80$). Bold values indicate that the value is optimal (i.e., reaches the theoretical upper bound $2n$).}
\label{tab:best_known_n3il_detailed}
\begin{tabular}{cccccccccccccccc}
\toprule
$n$ & 1 & 2 & 3 & 4 & 5 & 6 & 7 & 8 & 9 & 10 & 11 & 12 & 13 & 14 & 15 \\
\cmidrule{1-16}
$M(n)$ & \textbf{2} & \textbf{4} & \textbf{6} & \textbf{8} & \textbf{10} & \textbf{12} & \textbf{14} & \textbf{16} & \textbf{18} & \textbf{20} & \textbf{22} & \textbf{24} & \textbf{26} & \textbf{28} & \textbf{30} \\
\midrule
$n$ & 16 & 17 & 18 & 19 & 20 & 21 & 22 & 23 & 24 & 25 & 26 & 27 & 28 & 29 & 30 \\
\cmidrule{1-16}
$M(n)$ & \textbf{32} & \textbf{34} & \textbf{36} & \textbf{38} & \textbf{40} & \textbf{42} & \textbf{44} & \textbf{46} & \textbf{48} & \textbf{50} & \textbf{52} & \textbf{54} & \textbf{56} & \textbf{58} & \textbf{60} \\
\midrule
$n$ & 31 & 32 & 33 & 34 & 35 & 36 & 37 & 38 & 39 & 40 & 41 & 42 & 43 & 44 & 45 \\
\cmidrule{1-16}
$M(n)$ & \textbf{62} & \textbf{64} & \textbf{66} & \textbf{68} & \textbf{70} & \textbf{72} & \textbf{74} & \textbf{76} & \textbf{78} & \textbf{80} & \textbf{82} & \textbf{84} & \textbf{86} & \textbf{88} & \textbf{90} \\
\midrule
$n$ & 46 & 47 & 48 & 49 & 50 & 51 & 52 & 53 & 54 & 55 & 56 & 57 & 58 & 59 & 60 \\
\cmidrule{1-16}
$M(n)$ & \textbf{92} & \textbf{94} & \textbf{96} & \textbf{98} & \textbf{100} & \textbf{102} & \textbf{104} & \textbf{106} & \textbf{108} & \textbf{110} & \textbf{112} & \textbf{114} & \textbf{116} & \textbf{118} & \textbf{120} \\
\midrule
$n$ & 61 & 62 & 63 & 64 & 66 & 68 & 70 & 72 & 74 & 76 & 78 & 80 & & & \\
\cmidrule{1-16}
$M(n)$ & \textbf{122} & \textbf{124} & \textbf{126} & \textbf{128} & \textbf{132} & \textbf{136} & \textbf{140} & 140 & 140 & 140 & 144 & 148 & & & \\
\bottomrule
\end{tabular}
\end{table}

\paragraph{Previous Best-known Results of Smallest Complete Sets:}
\Cref{tab:optimal_smallest_complete_2_to_12} shows the sizes of the smallest complete sets for $2 \leq n \leq 12$. These results are obtained by exhaustive enumeration, so they are optimal \citep{AICHHOLZER2023101913}. \Cref{tab:current_smallest_complete_13_and_more} shows the best-known sizes of smallest complete sets for $n > 12.$ Note that the definition of ``independent dominating set'' in their work is equivalent to ``complete set'' in our context.

\begin{table}[htb]
\centering
\caption{Known optimal values for sizes of the smallest complete sets ($n \leq 12$).}
\label{tab:optimal_smallest_complete_2_to_12}
\begin{tabular}{cccccccccccc}
\toprule
$n$ & 2 & 3 & 4 & 5 & 6 & 7 & 8 & 9 & 10 & 11 & 12 \\
\midrule
$\mathscr{I}_n$ & 4 & 4 & 4 & 6 & 6 & 8 & 8 & 8 & 8 & 10 & 10 \\
\bottomrule
\end{tabular}
\end{table}

\begin{table}[htb]
\centering
\caption{Current best upper bounds for sizes of smallest complete sets ($n > 12$).}
\label{tab:current_smallest_complete_13_and_more}
\begin{tabular}{cccccccccccccccccccc}
\toprule
$n$ & 13 & 14 & 15 & 16 & 17 & 18 & 19 & 20 & 21 & 22 & 23 & 24 & 25 & 26 & 27 & 28 & 29 & 30 & 36 \\
\midrule
$\mathscr{I}_n \leq$ & 12 & 12 & 14 & 14 & 15 & 16 & 16 & 16 & 16 & 18 & 20 & 20 & 22 & 24 & 24 & 24 & 24 & 25 & 28 \\
\bottomrule
\end{tabular}
\end{table}

\section{Detailed Experiment Results}

\subsection{Ablation Study}

\begin{table}[H]
    \centering
    \caption{Incremental ablation study on $n\times n$ grids (Average terminal number of points and average runtime in seconds, with std in parentheses). Statistical significance vs.~the preceding variant is denoted by $^* p<0.05, ^{**} p<0.01, ^{***} p<0.001$. Search budget is $10n^2$ iterations.}
    \label{tab:incremental_ablation}
    \resizebox{\textwidth}{!}{
        \begin{tabular}{l c c c c}
            \toprule
            \multirow{2}{*}{\textbf{Algorithm Variant}} & \multicolumn{2}{c}{\boldmath$n=30$} & \multicolumn{2}{c}{\boldmath$n=40$} \\
            \cmidrule(lr){2-3} \cmidrule(lr){4-5}
             & \textbf{Points} & \textbf{Time} & \textbf{Points} & \textbf{Time} \\
            \midrule
            \textbf{M0}: Baseline ($\mathcal{A}_\text{naive}$) & $45.0\,(1.4)$ & $37.64\,(7.27)$ & $47.8\,(13.6)$ & $106.53\,(26.91)$ \\
            \hdashline
            \textbf{M1}: M0 + $\mathcal{A}_\text{feas}$ & $53.2\,(0.8)^{***}$ & $261.93\,(51.89)$ & $70.3\,(0.8)^{***}$ & $1533.96\,(250.77)$ \\
            \textbf{M2}: M1 + Incremental Update & $53.5\,(0.8)$ & $192.11\,(39.79)^{***}$ & $70.0\,(0.5)$ & $1134.80\,(241.91)^{***}$ \\
            \textbf{M3}: M2 + Canonical Pruning & $53.5\,(0.8)$ & $200.34\,(39.43)$ & $70.0\,(0.8)$ & $1184.56\,(288.49)$ \\
            \textbf{M4}: M3 + Batch Transitions ($C_4$) & $\mathbf{56.1}\,(0.4)^{***}$ & $46.34\,(10.06)^{***}$ & $72.0\,(0.6)^{***}$ & $282.59\,(97.78)^{***}$ \\
            \textbf{M5}: M4 + Anytime Optimal & $55.4\,(1.3)$ & $45.52\,(9.69)$ & $\mathbf{72.6}\,(1.1)^{*}$ & $277.33\,(106.73)$ \\
            \textbf{Ours}: M5 + Subtree Reuse & $54.9\,(1.7)$ & $\mathbf{41.41}\,(9.28)$ & $\mathbf{72.6}\,(1.3)$ & $\mathbf{250.42}\,(61.93)$ \\
            \bottomrule
        \end{tabular}
    }

    \vspace{2em}

    \resizebox{\textwidth}{!}{
        \begin{tabular}{l c c c c}
            \toprule
            \multirow{2}{*}{\textbf{Algorithm Variant}} & \multicolumn{2}{c}{\boldmath$n=50$} & \multicolumn{2}{c}{\boldmath$n=70$} \\
            \cmidrule(lr){2-3} \cmidrule(lr){4-5}
             & \textbf{Points} & \textbf{Time} & \textbf{Points} & \textbf{Time} \\
            \midrule
            \textbf{M0}: Baseline ($\mathcal{A}_\text{naive}$) & $45.0\,(1.0)$ & $201.67\,(4.96)$ & $64.8\,(0.8)$ & $1036.85\,(19.50)$ \\
            \hdashline
            \textbf{M1}: M0 + $\mathcal{A}_\text{feas}$ & $86.7\,(1.2)^{***}$ & $6421.70\,(1065.50)$ & $119.2\,(1.1)^{***}$ & $54667.12\,(1237.59)$ \\
            \textbf{M2}: M1 + Incremental Update & $86.2\,(0.9)$ & $3945.52\,(603.08)^{***}$ & $119.2\,(1.1)$ & $32974.75\,(676.13)^{***}$ \\
            \textbf{M3}: M2 + Canonical Pruning & $86.3\,(0.7)$ & $3868.81\,(456.44)$ & $119.2\,(1.0)$ & $33042.73\,(719.21)$ \\
            \textbf{M4}: M3 + Batch Transitions ($C_4$) & $\mathbf{90.2}\,(1.9)^{***}$ & $874.99\,(183.97)^{***}$ & $\mathbf{124.8}\,(2.2)^{***}$ & $7722.33\,(963.06)^{***}$ \\
            \textbf{M5}: M4 + Anytime Optimal & $89.8\,(1.4)$ & $940.59\,(167.67)$ & $124.6\,(1.7)$ & $7822.67\,(967.08)$ \\
            \textbf{Ours}: M5 + Subtree Reuse & $89.8\,(2.0)$ & $\mathbf{775.19}\,(128.46)^{***}$ & $124.3\,(1.5)$ & $\mathbf{6693.15}\,(599.49)^{***}$ \\
            \bottomrule
        \end{tabular}
    }
\end{table}
% Note: Ensure you have \usepackage{arydshln} in your preamble to render \hdashline correctly.

\clearpage
\section{Examples}
\label{appendix:examples}

To best demonstrate the scalability of our framework, we show examples of configurations discovered on larger grids for every problem evaluated in this study.

\subsection{Max No-Three-in-Line}
\begin{figure}[htbp]
    \centering
    \includegraphics[width=0.8\textwidth]{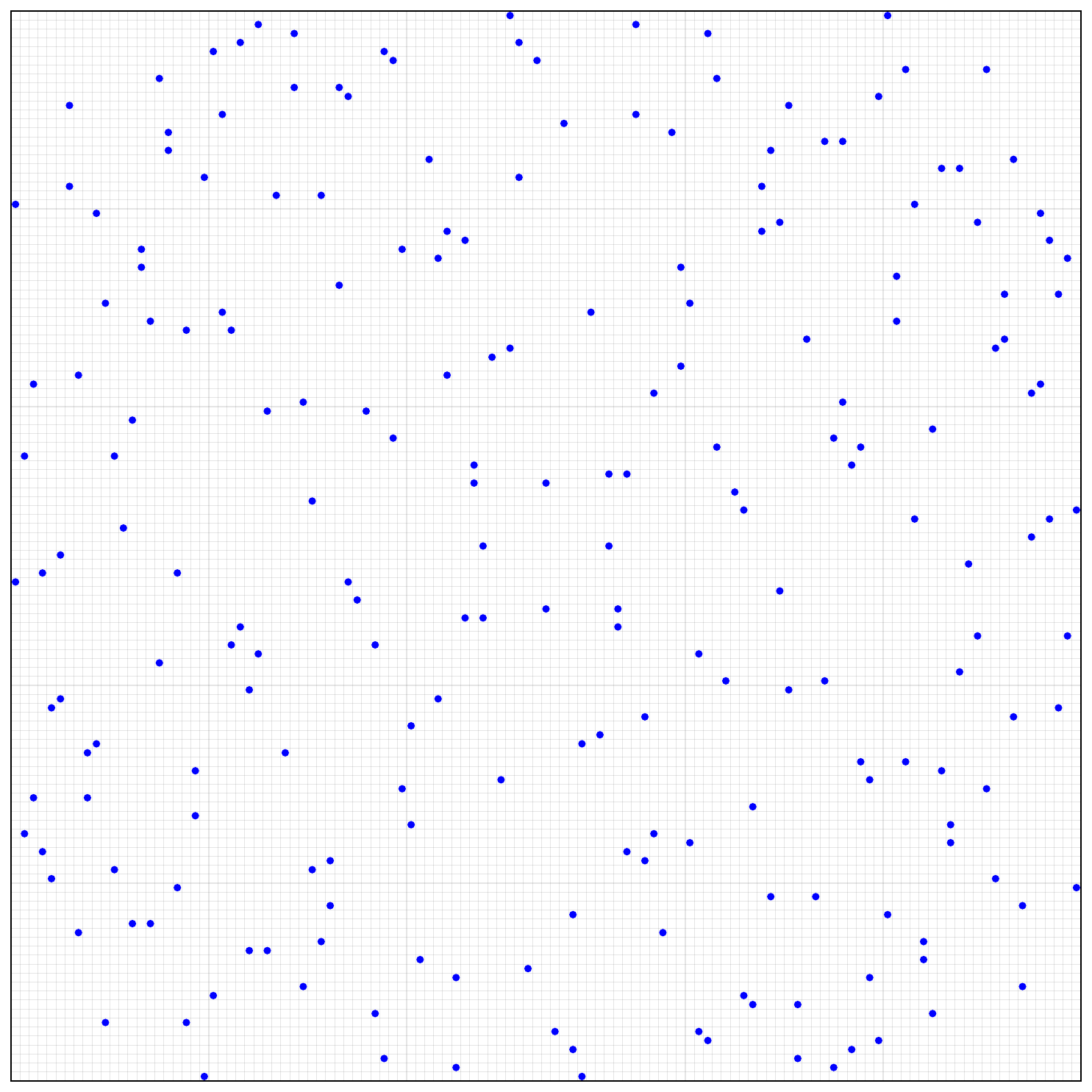}
    \caption{A discovered configuration with $216$ points on a $119 \times 119$ grid, satisfying the No-Three-in-Line constraint.}
    \label{fig:example_max_n3il}
\end{figure}

\clearpage
\subsection{Smallest Complete Set}
\begin{figure}[htbp!]
    \centering
    \begin{subfigure}[b]{\textwidth}
        \centering
        \includegraphics[width=0.5\linewidth]{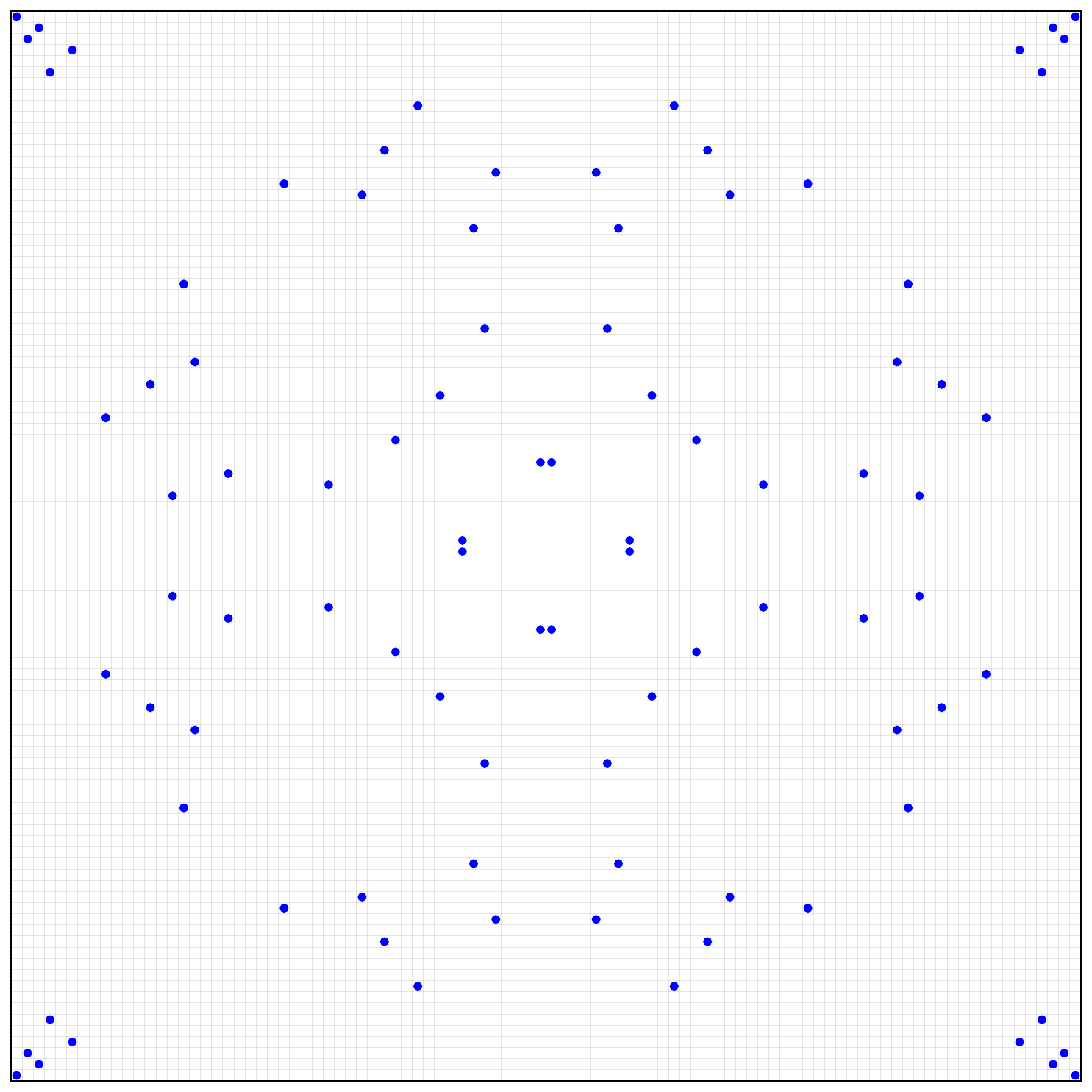}
        \caption{A complete set with $92$ points.}
        \label{fig:example_min_complete_points}
    \end{subfigure}

    \vspace{0.5em}

    \begin{subfigure}[b]{\textwidth}
        \centering
        \includegraphics[width=0.5\linewidth]{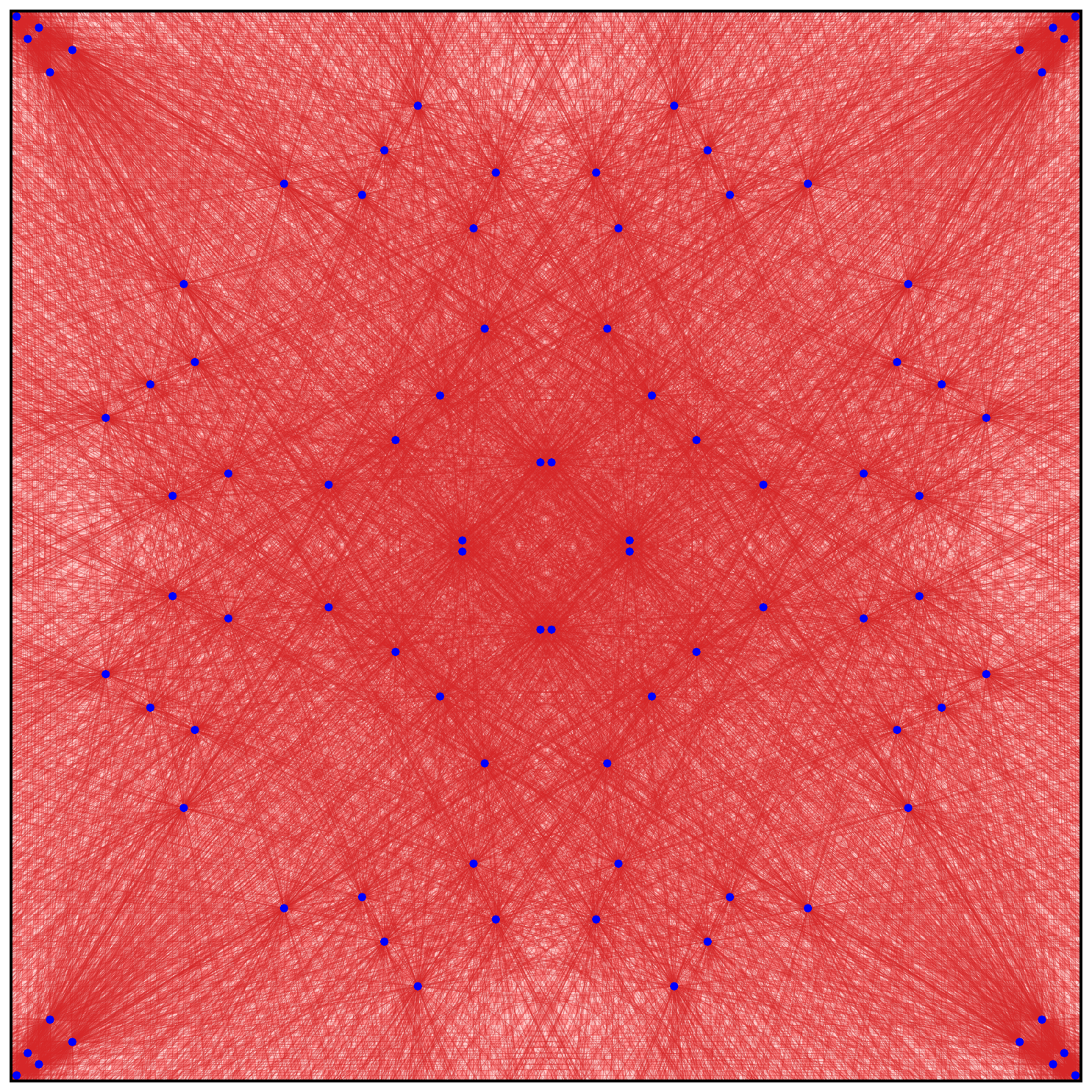}
        \caption{Raycast visualization demonstrating full grid coverage without violating collinearity constraints.}
        \label{fig:example_min_complete_coverage}
    \end{subfigure}

    \caption{A valid smallest complete set (independent geometric dominating set) discovered by our framework, consisting of $92$ points on $96 \times 96$ grid. The bottom panel visually confirms the validity of the complete set, illustrating how the rays generated by the points cover the entire grid while satisfying the No-Three-in-Line constraint.}
    \label{fig:example_min_complete_combined}
\end{figure}

\clearpage
\subsection{Smallest Geometric Dominating Set}
\begin{figure}[htbp!]
    \centering
    \begin{subfigure}[b]{0.6\textwidth}
        \centering
        \includegraphics[width=0.9\linewidth]{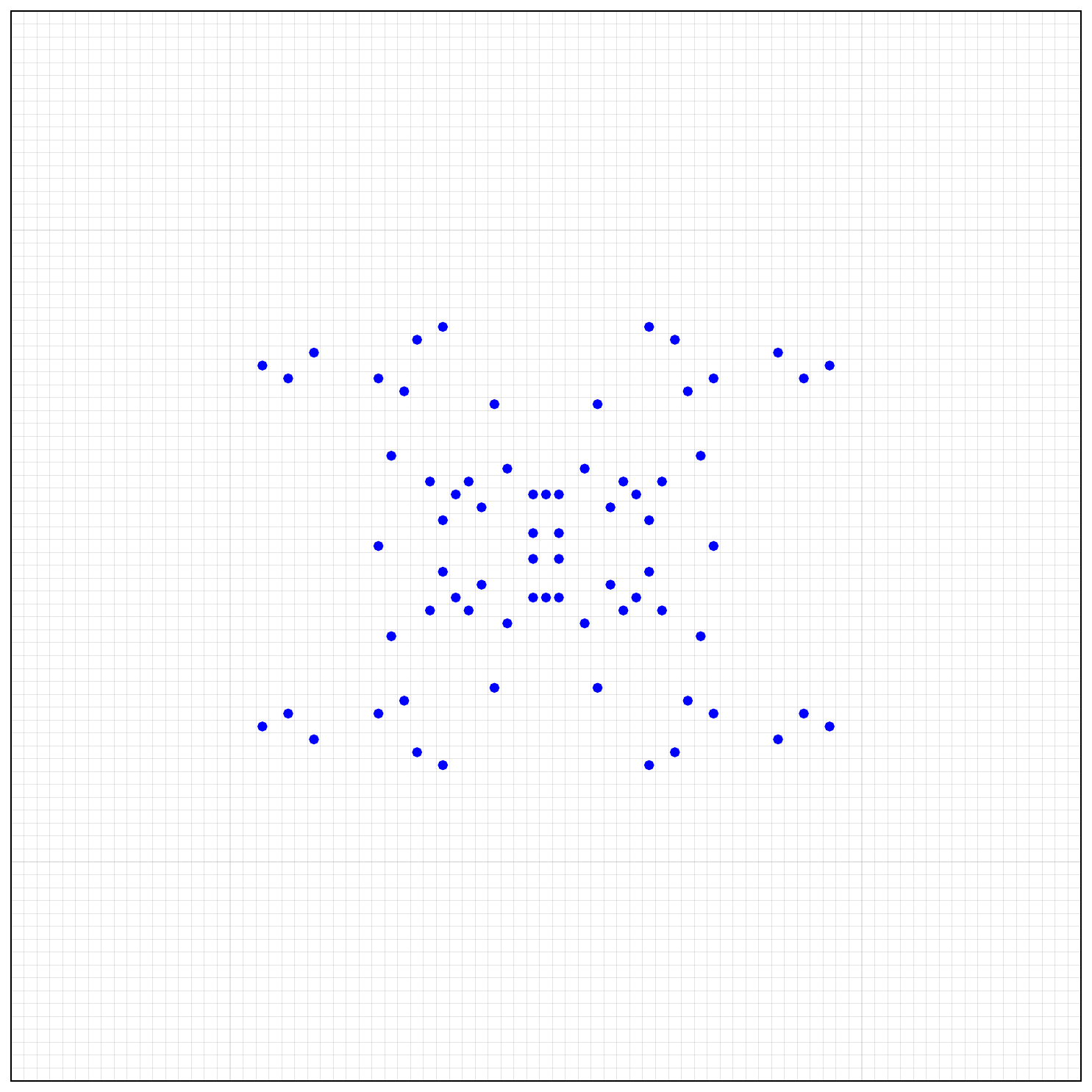}
        \caption{A geometric dominating set with $72$ points.}
        \label{fig:example_geodom_points}
    \end{subfigure}

    \vspace{0.5em}

    \begin{subfigure}[b]{0.6\textwidth}
        \centering
        \includegraphics[width=0.9\linewidth]{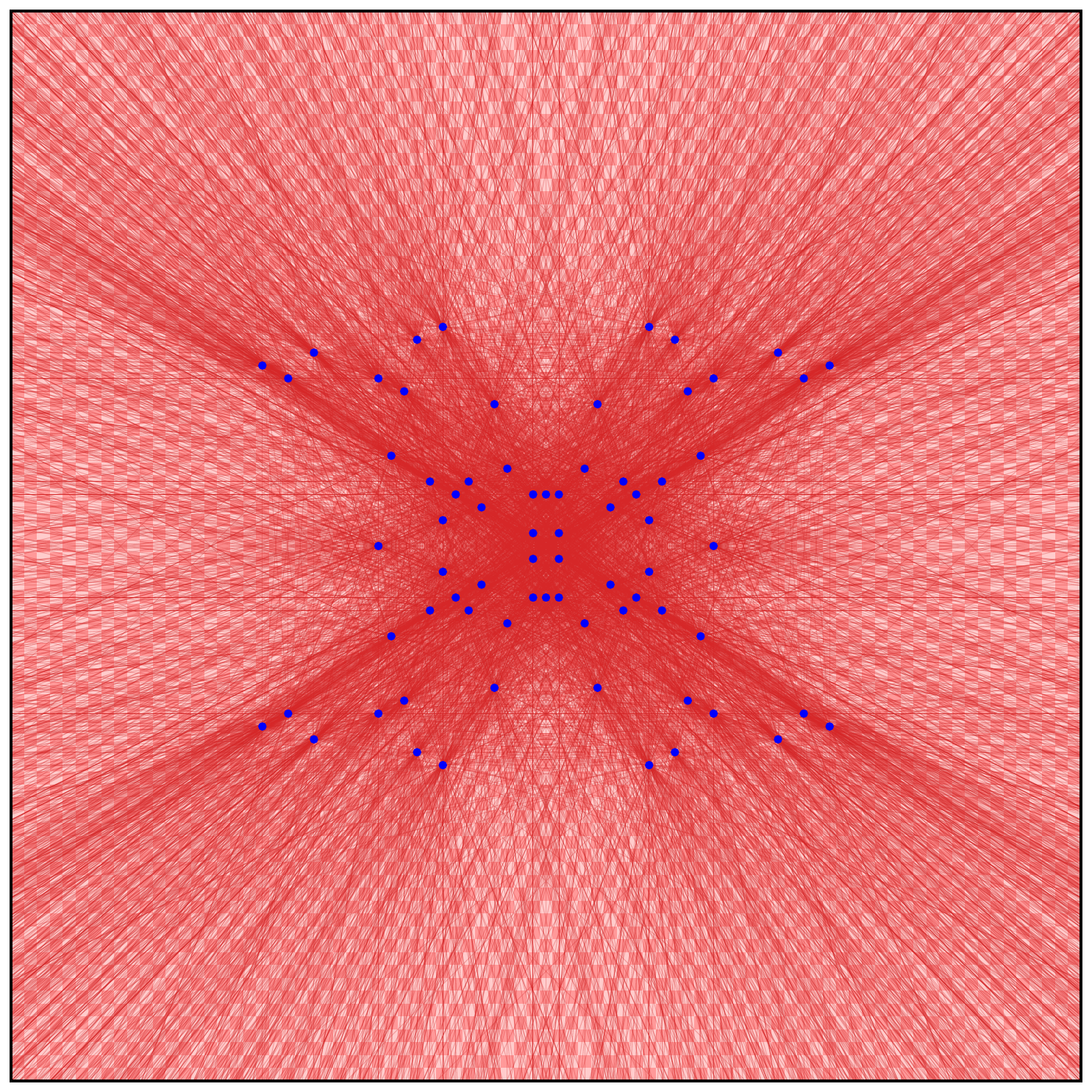}
        \caption{Raycast visualization demonstrating full grid coverage.}
        \label{fig:example_geodom_coverage}
    \end{subfigure}
    \caption{An example of a geometric dominating set on an $83 \times 83$ grid with $72$ points. The bottom panel visually confirms the validity of the dominating set, illustrating how the lines generated by the $72$ points completely cover the entire grid.}
    \label{fig:example_geodom_combined}
\end{figure}

\clearpage
\subsection{Max No-Four-in-Line}
\begin{figure}[htbp]
    \centering
    \includegraphics[width=0.8\textwidth]{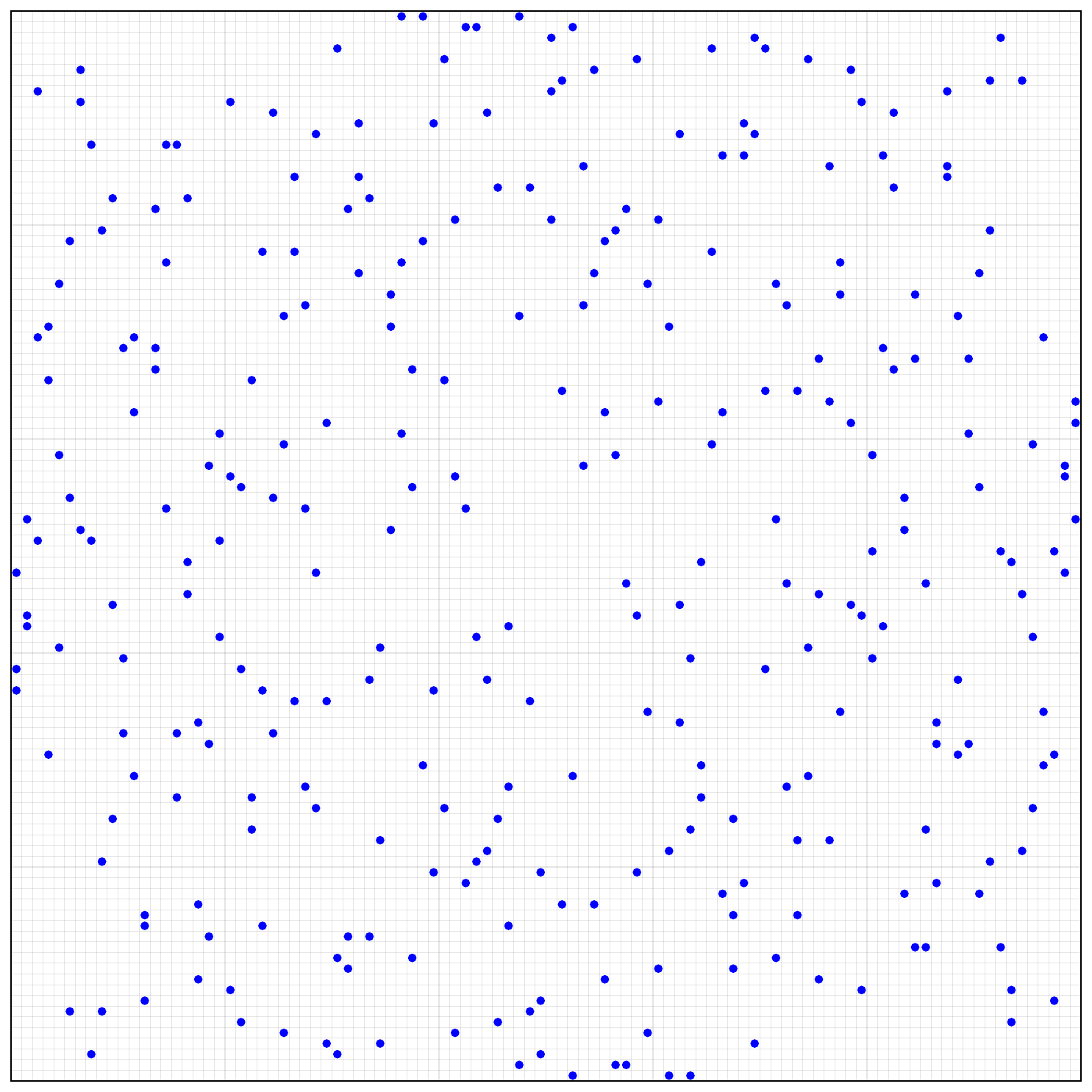}
    \caption{An optimal configuration with $300$ points on a $100 \times 100$ grid, satisfying the No-Four-in-Line constraint. }
    \label{fig:example_max_n4il}
\end{figure}

\clearpage
\subsection{Max No-Isosceles Triangle}
\begin{figure}[htbp]
    \centering
    \includegraphics[width=0.8\textwidth]{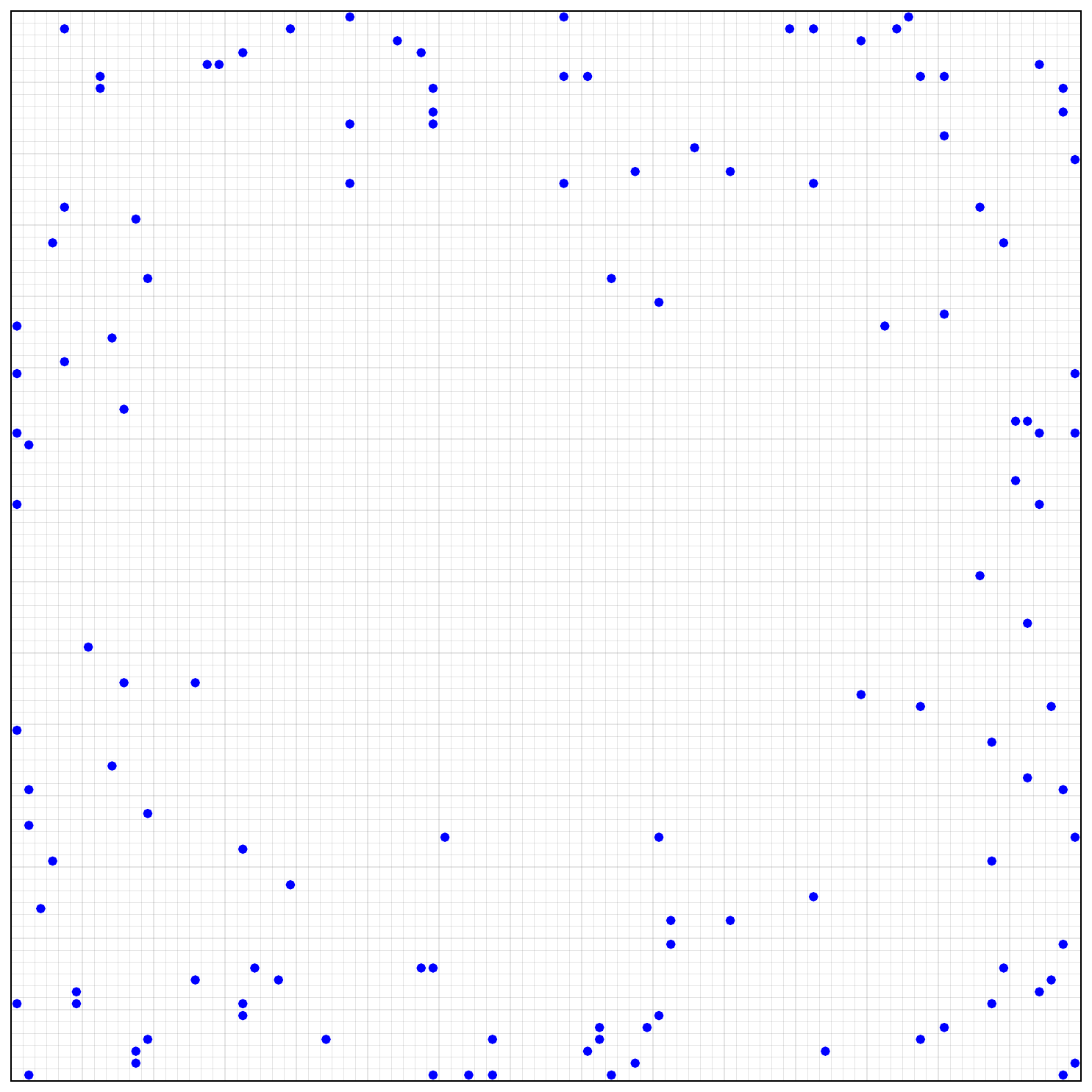}
    \caption{A configuration with $124$ points on a $90 \times 90$ grid, where no three points form an isosceles triangle.}
    \label{fig:example_no_isosceles}
\end{figure}

\clearpage
\subsection{Max No-Four-on-a-Circle}
\begin{figure}[htbp]
    \centering
    \includegraphics[width=0.8\textwidth]{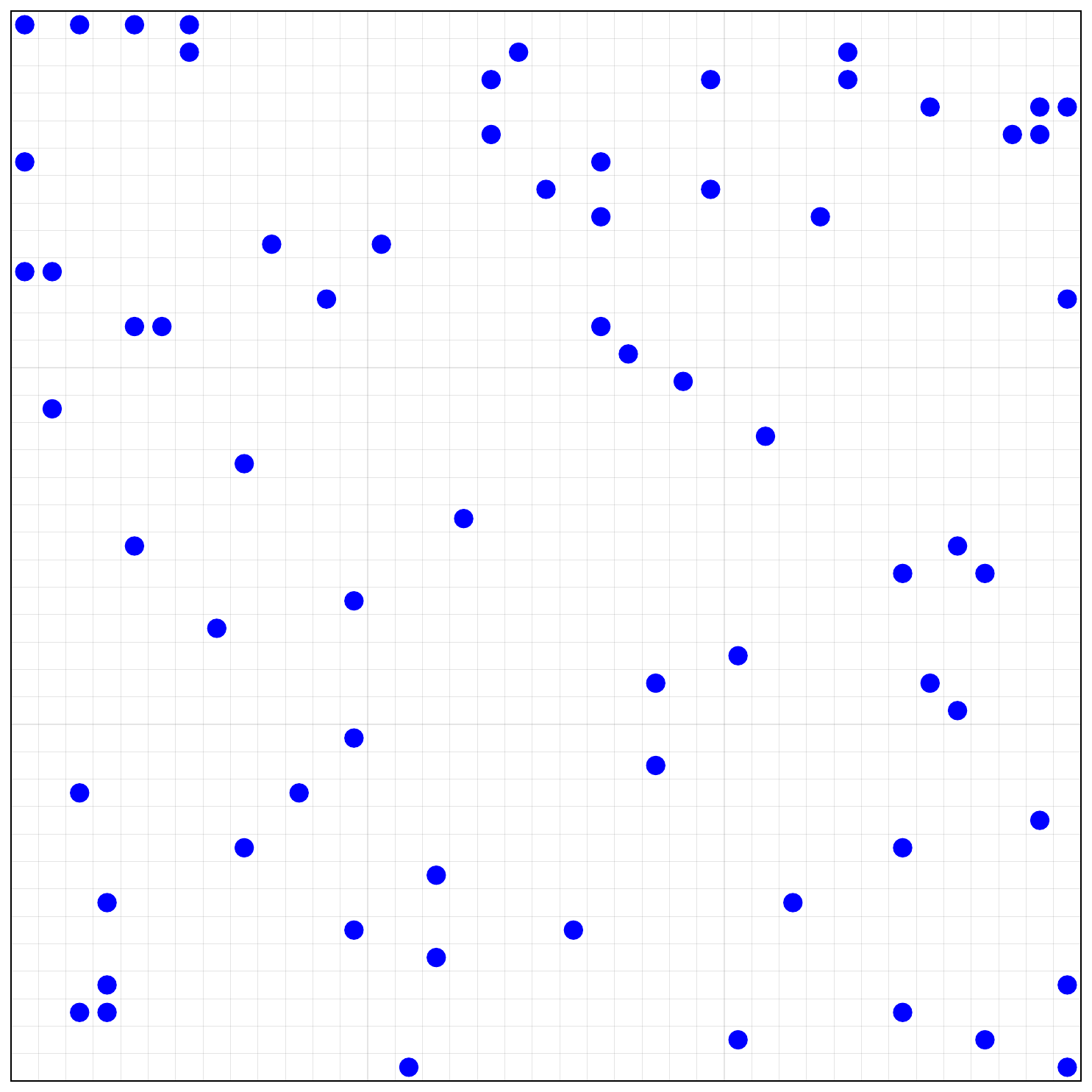}
    \caption{A configuration with $69$ points on a $39 \times 39$ grid, satisfying the No-Four-on-a-Circle constraint.}
    \label{fig:example_no_4_on_circle}
\end{figure}

\end{document}